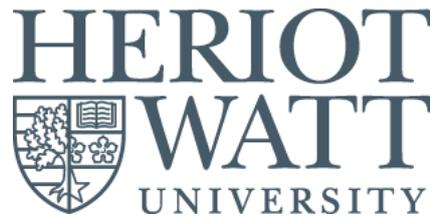

Master's Project

# Parkinson's Disease Diagnosis Using Deep Learning

Author:

Mohamad Alissa

| Supervisors: | Second Reader: |
|---|---|
| Dr Michael Lones | Dr Ron Petrick |
| Dr Marta Vallejo | |

August 2018
Department of Computer Science

School of Mathematical & Computer Sciences
Thesis submitted as part of the requirements for the award of the degree of M.Sc. in Artificial Intelligence with Speech and Multimodal Interaction

# Declaration

I, Mohamad Alissa,

confirm that this work submitted for assessment is my own and is expressed in my own words. Any uses made within it of the works of other authors in any form (e.g., ideas, equations, figures, text, tables, programs) are properly acknowledged at any point of their use. A list of the references employed is included.

Signed: Mohamad Alissa

Date:   16th August 2018

# Abstract

Parkinson's Disease (PD) is a chronic, degenerative disorder which leads to a range of motor and cognitive symptoms. PD diagnosis is a challenging task since its symptoms are very similar to other diseases such as normal ageing and essential tremor. Much research has been applied to diagnosing this disease. This project aims to automate the PD diagnosis process using deep learning, Recursive Neural Networks (RNN) and Convolutional Neural Networks (CNN), to differentiate between healthy and PD patients. Besides that, since different datasets may capture different aspects of this disease, this project aims to explore which PD test is more effective in the discrimination process by analysing different imaging and movement datasets (notably cube and spiral pentagon datasets). In addition, this project evaluates which dataset type, imaging or time series, is more effective in diagnosing PD.

## Acknowledgements

I would like to thank Dr Michael Lones and Dr Marta Vallejo for their helping, advice, guidance and their time throughout the project. and I would like to thank the second reader, Dr Ron Petrick, for the valuable feedback provided.

## Table of Contents

# Figures List

# Tables List

# Abbreviations List

| | |
|---|---|
| **BPTT** | Backpropagation Through Time |
| **BRNN** | Bidirectional Recurrent Neural Networks |
| **DL** | Deep Learning |
| **EDS** | Excessive Daytime Sleepiness |
| **FA** | Firefly Algorithm |
| **FN** | False Negative |
| **FP** | False Positive |
| **HSD** | Honestly Significant Difference |
| **LRCN** | Long-term Recurrent Convolutional Network |
| **LSTM** | Long Short-Term Memory |
| **MCI** | Mild Cognitive Impairment |
| **ML** | Machine Learning |
| **NN** | Neural Network |
| **PD** | Parkinson's Disease |
| **PDD** | PD Dementia |
| **PD-MCI** | Mild cognitive impairment |
| **PSO** | Particle Swarm Optimisation |
| **RA** | Random Search |
| **RBD** | REM sleep Behaviour Disorder |
| **ReLU** | Rectified Linear Unit |
| **TBPTT** | Truncated Backpropagation Through Time |
| **TN** | True Negative |
| **TP** | True Positive |
| **UPDRS** | Unified Parkinson's Disease Rating Scale, |

# Chapter 1. Introduction

## 1.1 Motivation

The brain is the main controller of our body. Therefore, any damage to this sensitive part of the human body will affect badly on the other organs. One of these negative effects is Parkinson's disease. As Pereira et al. (2017) note, Parkinson's disease (PD) is a chronic, progressive, neurodegenerative disorder which begins when a certain area of the brain has been damaged. PD symptoms overlap with other diseases like normal ageing and essential tremor, especially in the early stages of these diseases (Samii et al., 2004). Thus, it is important to differentiate between PD and other diseases in order to give the right treatment to the patient.

Mathematical models such as Deep Learning (DL) provide a suitable technique to detect disease symptoms (Pereira et al., 2017). These modelling approaches include topologies specialized for some kind of datasets such as imaging datasets and time-series datasets (i.e. a dataset comprised of a set of sequences, where each sequence contains data points that are indexed in time order). Therefore, it is worthy to investigate the deep learning techniques on the PD, especially after the recent success of deep learning in different fields.

## 1.2 Objectives

In general, the main aim of this thesis is to automate the PD diagnosis process in order to discover this disease as early as possible. If we discover this disease earlier, then the treatments are more likely to improve the quality life of the patients and their families (Pereira et al., 2017).

Much research has taken place in this field, but the aim of this thesis is to use different deep learning techniques to detect PD accurately by analysing the patients' motor features and hand-drawn images.

The following tasks are required to achieve the objectives mentioned above:-

1- Develop predictive models to differentiate between healthy people and people with PD.
2- Investigate which time series datasets are more useful for training predictive models.
3- Comparing two different imaging datasets (each one related to a different PD drawing test) to examine which one is more useful for training predictive models.
4- Explore whether imaging datasets or time series (movement) datasets are more effective as a basis for discrimination.
5- Study and analyse different deep learning models, including CNNs and RNNs.

This report is organised as follows:- Literature review chapter 2 including Parkinson's disease section 2.1; machine learning section 2.2; deep learning section 2.3; some related works section 2.4 and PD diagnosis problems section 2.5, the datasets chapter 3, the methodology chapter 4, the implementation chapter 5, Discussion & Evaluation chapter 6 and the conclusion & future work chapter 7.

# Chapter 2. Literature Review

This part of the dissertation describes the theoretical background of this project, starting with an explanation of Parkinson's disease, followed by overviews of machine learning, deep learning, related work and finally PD diagnosis problems.

## 2.1 Parkinson's disease (PD)

### 2.1.1 Overview

PD is one of the diseases that happens when cells stop working properly or damage happens in a part of the brain called the substantia nigra pars compacta. The cells in this region are responsible for producing a very important chemical substance called dopamine. These cells are lost over the time, so the brain loses the dopamine which is responsible for controlling the movement behaviours in the body including walking, writing, and even smiling (Vallejo et al., 2016).

In general, as Pereira et al. (2017) reported, this disease is defined as a chronic, progressive neurodegenerative disorder. PD afflicts humans around the world, especially in countries with a high average age of the population (Vallejo et al., 2016). According to Parkinson's diseases foundation (2015), about 10 million people worldwide have PD, one million of them in the United States (Parkinson's Disease Foundation, 2015). The website of the Parkinson's Disease Society stated that one individual in every 500 British people has this disease and it is expected that this number will increase 3-fold in the next 50 years. Normally this illness becomes worse over time and mostly affects people between 50-70 years old. PD was first described by James Parkinson, a British physician, in 1817 and there is still no treatment for PD ( Lones et al., 2014; Pereira et al., 2015).

**2.1.2 PD Types & Symptoms**

PD mainly includes motor symptoms (a movement disorder) and non-motor symptoms like cognitive dysfunction (Meireles et al., 2012).

For motor symptoms, four main signs are considered as cardinal symptoms: rest tremor, rigidity, bradykinesia, and sometimes postural instability. About 70% of PD patients have a resting tremor which is between 3-5 HZ and it characterized as asymmetrical tremor. The second sign of PD is a feeling of resistance during joints' movements and it is called cogwheel rigidity (Samii et al., 2004). In other words, it is the converse of smooth movements (Khan Academy, 2015). Slowing down the movement is the third sign, called bradykinesia; it enlarges with simple movements like handwriting. The fourth symptom is postural instability and this one does not happen in the early stage of PD, in particular for younger patients and it is related to balance, which makes the patient unstable on their feet and may lead to falls (Samii et al., 2004; Khan Academy, 2015).

Some of the non-motor symptoms of PD, like hyposmia, rapid eye movement (REM), sleep behaviour disorder, constipation, and depression may emerge before any motor symptoms by years (Meireles et al., 2012). Many patients also exhibit cognitive dysfunction, and these range from what is called mild cognitive impairment (PD-MCI) to PD dementia (PDD) (Litvan et al., 2012). In several cases, PD-MCI emerges in the early stage of the disease, while PDD tends to occur after 20 years of having the PD. PD-MCI is defined as thinking and memory problems abnormal to what is expected with normal ageing, but without preventing the patient from carrying out daily routine activities. Moreover, PD-MCI diagnosis is important because it could be a transition to PDD (Meireles et al., 2012). PDD symptoms include impaired short-term memory, executive dysfunction, attention impairment, visual-spatial deficit, behavioural or neuropsychiatric symptoms like psychotic symptoms (hallucinations), changes in personality and mood, anxiety and apathy (Meireles et al., 2012). Figure 1 shows how the symptoms develop over time until reaching the dementia stage.

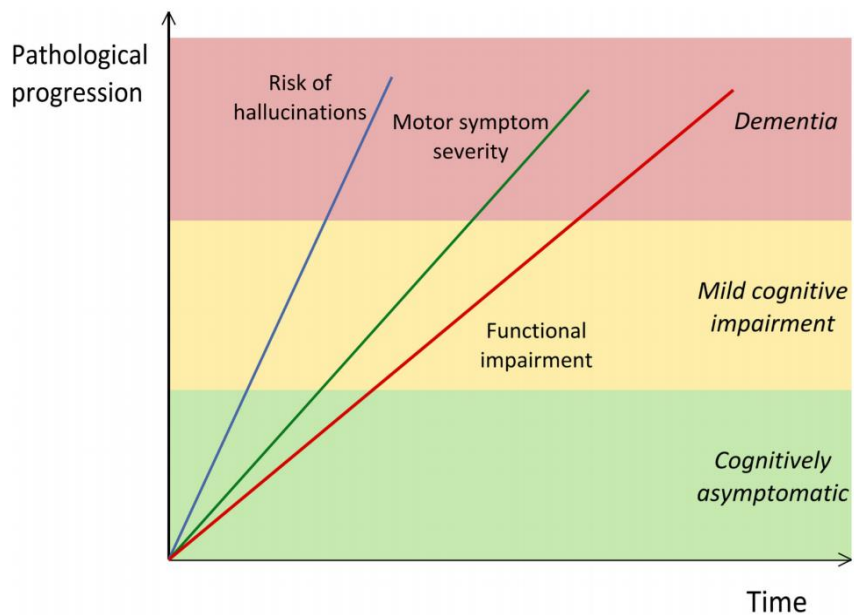

*Figure 1* PD symptoms development over time (Meireles et al., 2012)

**2.1.3 Parkinson's disease diagnosis**

First of all, it is important to differentiate between PD and parkinsonism. Parkinsonism is the clinical presentation for a set of symptoms which could be attributed to PD such as tremors, rigidity, and bradykinesia. PD is considered as one form of parkinsonism, so not all the subjects present the parkinsonism symptoms will have PD as it could be due to other health issues such as a vascular disorder or other neurodegenerative diseases (Parkinson's Disease Foundation, 2015).

There is difficulty to distinguish PD from the other types of parkinsonism by only monitoring the motor features of PD patients. Therefore, it is necessary to add other features to increase the clinical diagnostic accuracy, like asymmetry and a robust response to levodopa treatment (Berg et al., 2013).

There are several obstacles facing the diagnosis criteria from motor features including:-

1- This criterion focuses on movement problems but, as mentioned in section 2.1.2, PD is also normally associated with non-motor features which do not respond to levodopa treatment and occur before any motor symptoms (Berg et al., 2013).

2- A main measure in the criteria is the response to dopamine replacement, which also happens in other types of parkinsonism (Berg et al., 2013).

The UK Parkinson's Disease Society Brain Bank Diagnostic Criteria comes as an update of the earlier diagnosis and includes both asymmetry and a robust response to levodopa treatment to increase the specificity of the diagnosis. Also, the diagnostic specificity has raised from what early studies suggested was 76% to 90% specificity at death, as follow-up studies reported (Berg et al., 2013).

Figure 2 illustrates the degree of the disability over time. It can be seen that PD diagnosis happens when the motor features start to enlarge (time = 0), but as this thesis mentioned in section 2.1.2 and 2.1.3, non-motor symptoms can precede the diagnosis by 20 years or more. Additional non-motor symptoms emerge after the PD diagnosis until reaching dementia, as the author mentioned in section 2.1.2. Also, It can be seen that postural instability tends to occur in advanced stages of the disease.

EDS=excessive daytime sleepiness. MCI=mild cognitive impairment. RBD=REM sleep behaviour disorder (Kalia et al., 2015). UPDRS is the Unified Parkinson's Disease Rating Scale, this measurement gives a score, 5-point scale, for each symptom (Goetz et al., 2008).

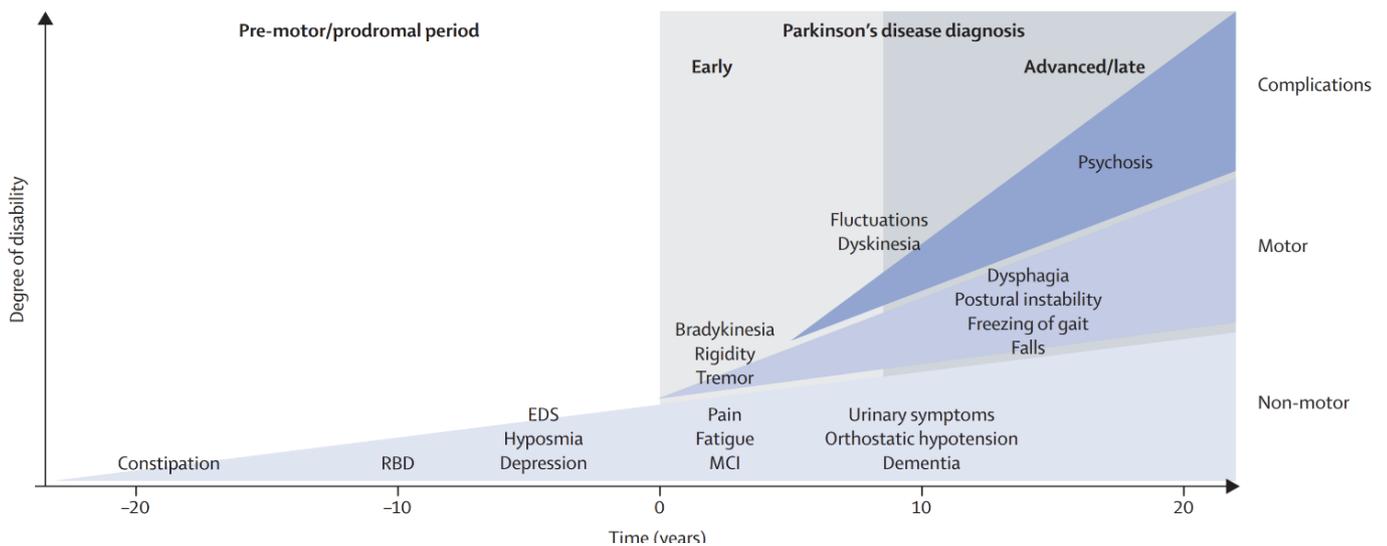

*Figure 2* degree of the disability over time (Kalia et al., 2015)

## 2.2 Machine learning

Machine learning (ML) is an important area in computer science. ML, called sometimes automated learning, is a collection of algorithms which aim to make computers learn from available input (called training data or representing experience) and give us the output as expertise (Shalev-Shwartz et al., 2014). Generally, We use machine learning in automating several tasks such as tasks performed by humans and tasks are over human capabilities.

### 2.2.1 Types of learning

Learning, according to Shalev-Shwartz et al. (2014), can be defined as "using the experience to gain expertise". According to Haykin (2009), there are two main types of learning:-

1. Supervised learning (learning with a teacher): in this type of learning we have input-output pairs. The teacher is able to give the algorithm the "optimum" response (output) for each input, and the algorithm (student) gives the actual output. Based on the differential between both optimum and actual output then the algorithm will change its parameters to reduce the error. So, the student learns from the knowledge provided by the teacher. This type of learning is normally related with classification task, which is the process of teaching a classifier the relationship between the model's input and output then use this expertise later for un-seen input (Witten et al., 2011).

2. Unsupervised learning (learning without a teacher):- in this kind of learning we do not have a teacher. As a consequence, we do not have a feedback to learn from, as in supervised learning. In this kind of learning, the algorithm learns how to represent the input instead of carrying out a prediction process. Under this type of learning, we have reinforcement learning is specialized in interaction problems like playing a game. In this type of learning the student learns from his mistakes during interaction with the environment (Sutton and Barto, 1998). This type of learning is normally related with Clustering which is the process of grouping the model's input to clusters in absence of the output (Witten et al., 2011).

### 2.2.2 Neural network (NN)

According to Shalev-Shwartz et al. (2014), a neural network is a directed graph containing several nodes which connect to each other by links, and the weighted sum of the output of the node passes as an input to the next nodes. These networks are inspired by the human brain (Haykin, 2009).

NN is a supervised ML algorithm which belongs to the classification task. NN can be divided into several types but we can group these types based on the network structure as follows.

1. Single-Layer Feedforward Networks: it is a simple form of NN and contains an input layer connected directly to an output layer.

2. Multilayer Feedforward Networks: this type differs from the previous type by the presence of hidden layers (at least one layer) between the input and output layers. This thesis will describe this type in section 2.3.2.

3. Recurrent Networks: this type differs from the feedforward networks by the presence of one or more feedback loops from the output of the neurons to the other neurons. This thesis will describe this type in section 2.3.3.

## 2.3 Deep learning (DL)

### 2.3.1 Overview

Deep learning is a relatively new approach within the field of neural networks. It is a branch of machine learning which deals with unstructured data (hierarchy data) like audio, image and text (Wang et al., 2017). DL represents the data in several layers with several levels of abstraction (Lin et al., 2017).

**Why Deep learning?**

The limitation of the shallow neural network (a standard NN with two hidden layers as maximum) leads to the necessity of depth. For example, in shallow NN, the number of neurons in the hidden layers grows exponentially with task complexity. So the shallow NN would need more neurons than the deep one (Goodfellow et al., 2016). Also, the human nervous system is a deep system (Wang et al., 2017).

**Deep learning applications**

Several prominent applications recently used DL, such as voice recognition, face recognition, prediction and machine translation(Wang et al., 2017). DL has been applied successfully in the medical field such as cancer detection (Cruz-Roa et al., 2013) and PD diagnosis (Pereira et al., 2016; Pereira et al., 2017).

DL architectures can be divided into several topologies including convolutional neural network (CNN) and recursive neural network (RNN) (Wang et al., 2017).

### 2.3.2 Convolutional neural network (CNN)

### 2.3.2.1 CNN Definition

As Wang et al. (2017) reported, a CNN is a feedforward network developed from the human visual cortex. Besides that, CNN can deal with huge amount of data which improves its ability to learn from this data. Also, CNN uses a convolution multiplication in its layers, with at least one layer. This makes this network specialized in processing data that have grid-like topology, like image data (Lin et al., 2017). CNN consists of several convolution layers, with each layer including three cardinal stages:-

1-convolution

 2-nonlinear activation (non-linearity transform), and

 3-pooling (sub-sampling).

In some literature, convolution and nonlinear activation are considered as the convolution layer and pooling (sub-sampling) is considered the second layer (Wang et al., 2017). Sometimes batch normalisation (BN) layers are used to improve the performance of a CNN. These layers are normally inserted before the non-linear activation layer.

This thesis follows the first interpretation, that is, 1-convolution, 2-nonlinear activation, and 3-pooling(sub-sampling).

### 2.3.2.2 CNN layers

**1-Convolution layer**

Convolution layer is the layer which applies the convolution operation (Wang et al., 2017). The convolution operation, in mathematics, is an operation on two functions as we see in the following equation (1):

$$h(t) = \int_{-\infty}^{+\infty} f(\tau)g(t-\tau)d\tau \qquad \text{Equation 1}$$

In CNN terminology ( f ) is the input, an input image for example, and ( g ) is called the filter (or kernel). However, according to Lin et al. (2017), we can find equation 2 in machine learning libraries which is equivalent to equation 1 after some mathematical operations.

$$h(t) = \sum_{\tau=1}^{vk} f(t+\tau)g(\tau) \qquad \text{Equation 2}$$

Where vk is the maximum valid indices of a kernel function (Lin et al., 2017).

**2- Nonlinear activation layer or non-linearity transform**

In the past, a sigmoid function, equation 3, had been used due to its good features like simple to find the derivatives and produce an output between [0, 1] as shown in figure 3a. however, the sigmoid function has a dangerous property that this function has a flat line when the output is very close to one or zero which squashes the gradients. This is known as the vanishing gradient problem. This property onsets significantly when the neural network becomes deeper which make the neural network more difficult to train (Lin et al., 2017).

$$\sigma(x) = 1/(1 + e^{-x}) \qquad \text{Equation 3}$$

Rectified linear unit (ReLU) is a solution to this problem. This function gives zero if the input is negative and keeps the input if the input is positive and has this simple form (Lin et al., 2017).

$$f(x) = \max(0,x) \qquad \text{Equation 4}$$

This function brings several advantages including:- it keeps the gradients on the right side of the function leading to faster training than using the sigmoid function. And it is more computationally efficient to compute than sigmoid function since ReLU does not include the exponential function. However, the left side of ReLU is always zero and this may lead to killing the gradients on this side, followed by a failure in the training process (Lin et al., 2017). The solution to this problem was to use leaky rectified linear units (leaky ReLU) which is a modification of ReLU as we see in figure 3c. It has this form:

$$f(x) = \begin{cases} x & \text{if } x > 0 \\ \alpha x & \text{otherwise} \end{cases}$$

An alpha of 0.01 is frequently used in machine learning applications, but alpha = 0.1 is applied in the figure 3c for clarification purposes (Lin et al., 2017). In general, this layer provides, after the convolution layer, benefits to the extraction of good features and makes the CNN closer to visual cortex functions (Wang et al., 2017).

Figure 3 shows the differences between sigmoid figure 3a, rectified linear units ReLU figure 3b and leaky ReLU figure 3c.

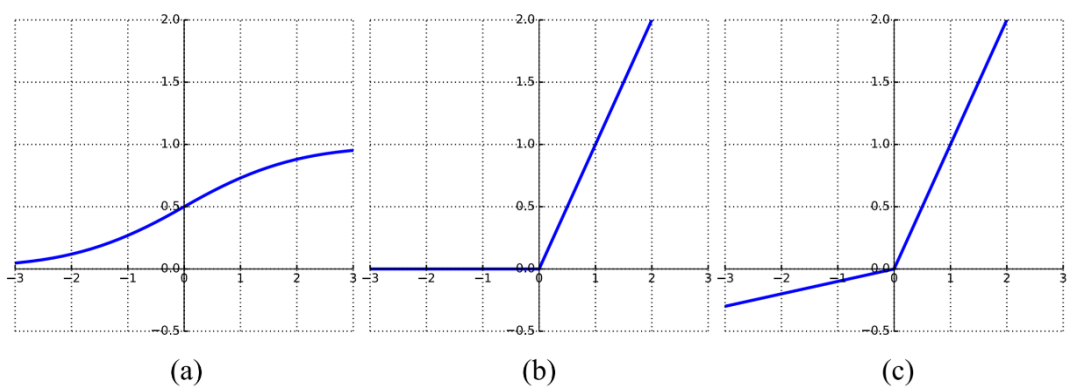

(a)      (b)      (c)

**Figure 3** *sigmoid (3a), rectified linear units ReLU (3b) and leaky ReLU (3c) functions (Lin et al., 2017)*

### 3- Pooling (sub-sampling) layer

This layer's task is to pick out one value from a set of values in the region it looks into (Wang et al., 2017). It has two properties: kernel (filter) size and the stride. The kernel size is the size of the region that the pooling layer looks into, and the stride is the difference of movements between two followed steps (Lin et al., 2017).

There are different ways to pick up the new values including max pooling to choose the maximum from the region, and the average pooling to retrieve the average value of all the inputs from the region. There is also a probabilistic pooling to pick a random value from the region (Wang et al., 2017).

The pooling layer brings several benefits to CNN, such as that small changes in the input will not affect the output of pooling when we use max or average pooling. Thus, we keep the CNN invariant to small distortions in the input image (Wang et al., 2017). In addition, this layer reduces the input size which will provide improvement and savings of complexity of the computation (Lin et al., 2017).

### 2.3.2.3 CNN Training

Although the progress of the deep neural network has increased widely in recent years, the training process in this kind of networks is still difficult and is an open area of contemporary research (Kohlbrenner et al. 2017). The training process in the neural network means to find the layers' weights which achieve the highest results in its work, e.g. classification. There are several approaches that have been applied to train CNNs. In general, as this thesis mentioned, a CNN consists of several layers. The first one is called the convolutional layer. This layer, in practical words, works as a feature extractor and the next layers work as a classifier (Kohlbrenner et al., 2017). So we have two sets of weights which we can consider regarding the training process, the convolutional layer weights and the classification (next layers) weights.

For the first weight set (convolutional layer weights), these weights might be initialized either randomly or pre-trained by using techniques like a convolutional auto-encoder (CAE) (Kohlbrenner et al., 2017). The second weights set (weights of layers coming after the convolutional layer, i.e. classification weights) can be initialized randomly. Both of these weights sets can be tuned using back-propagation, i.e. we can tune convolutional and classification weights to minimize the error calculated by using the objective function. This error is the difference between the predicted class and the true class (Lin et al., 2017).

Convolutional Auto-Encoder (CAE) is a special kind of CNN. The difference between them is that the CNN learns how to filter and combine the features in order to classify the input, while the CAE learns how to filter in order to extract features which can be used to reconstruct the input. CNNs are usually related to supervised learning, while the CAE relates to unsupervised learning (Galeone, 2016). Using a CAE with a CNN is considered to be a way of optimizing the CNN (Kohlbrenner et al., 2017).

Fig 4a shows an example of using CAE with CNN and figure 4b shows back-propagation to tune the classification weights.

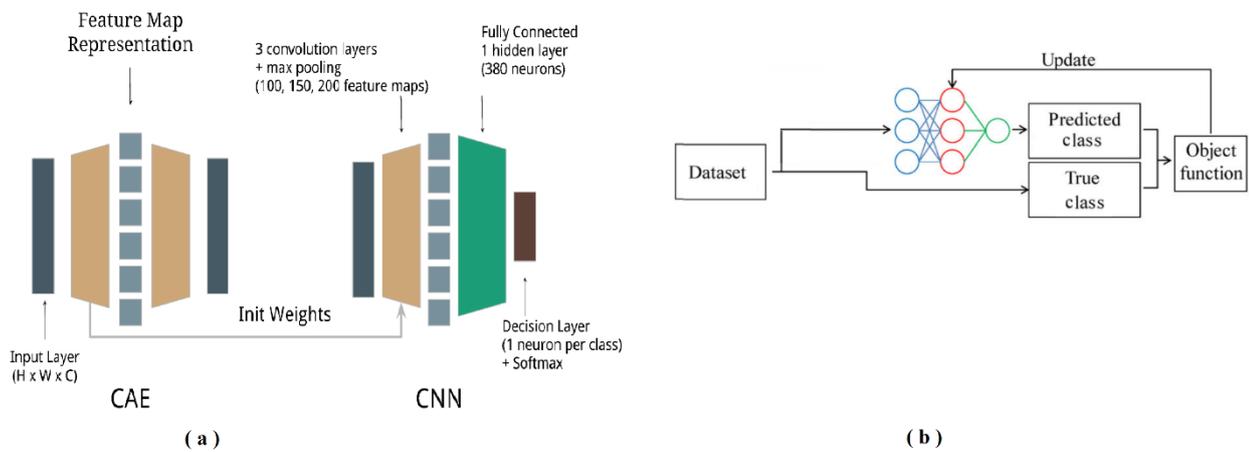

**Figure 4** *using CAE with CNN (4a), using back-propagation to tune the classification weights (4b)*

*(Kohlbrenner et al.; Lin et al., 2017)*

## 2.3.3 Recurrent neural network (RNN)

### 2.3.3.1 RNN definition

RNN is a kind of neural network which is adapted for sequential data like time series data (Wang et al., 2017). The main idea of RNN is the presence of cyclic connections inside the network, and this is the difference between feedforward networks like CNN and RNNs (Graves, 2014). This idea has led to several types of RNN including Jordan networks, Elman networks, time delay neural networks, and echo state networks (Graves, 2014).

### 2.3.3.2 Early recurrent network designs
According to Lipton et al. (2015), Jordan and Elman networks are considered as early forms of RNN.

The Jordan network (Fig 5), introduced in 1986, is a feedforward network with one hidden layer and a special unite extension (Lipton et al., 2015). According to (Wang et al., 2017), this network can be formulated as:-

$$h^t = \sigma(W_h X + W_r y^{t-1})$$
$$y = \sigma(W_y h^t)$$

Where X is the input vector, Wh are the weights of the hidden layer, Wr are the weights of the recurrent computation, Wy are the weights of the output layer, h refers to a hidden layer, y refers to output, t refers to time step (Wang et al., 2017).

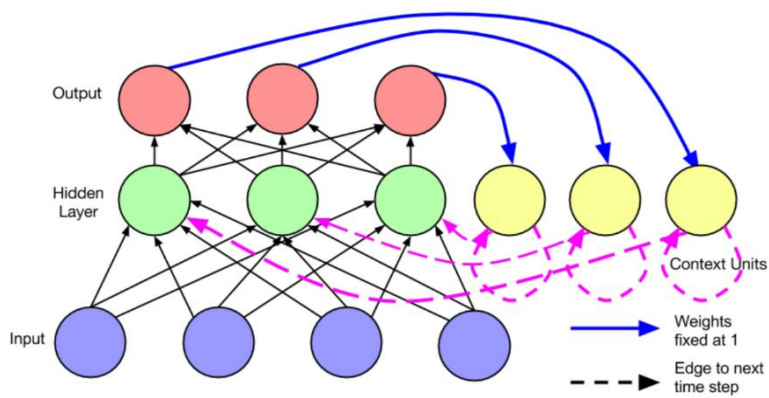

*Figure 5* Jordan Network Structure (Lipton et al., 2015)

In 1990, Elman introduced a new RNN (Fig 6), which is slightly different to a Jordan network (Fig 5). According to Wang et al. (2017), his model can be formulated as-

$$h^t = \sigma(W_h X + W_r h^{t-1})$$
$$y = \sigma(W_y h^t)$$

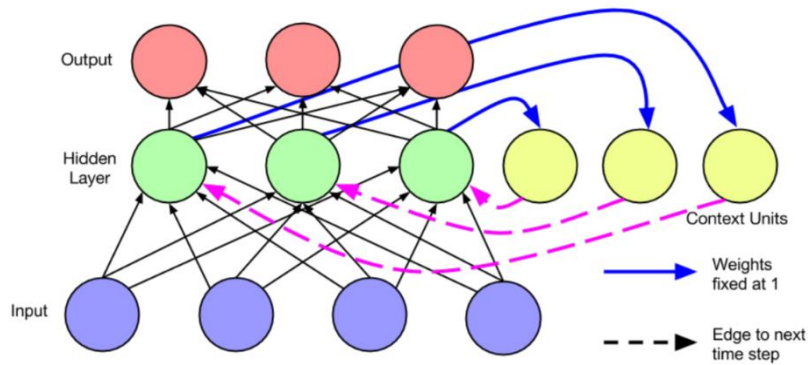

***Figure 6*** *Elman Network Structure (Lipton et al., 2015)*

As the figures 5 & 6 present, the difference between Jordan and Elman is that in a Jordan network the information of the previous time step is obtained from the previous output, while in Elman network this information is provided from the previous hidden layer (Wang et al., 2017).

**2.3.3.3 Training recurrent networks**

RNN structure prevents traditional backpropagation being applied since with RNN there is no stopping point where the backpropagation can stop. The solution was to unfold the RNN structure, which means expanding the RNN to several neural networks with certain time steps and applying traditional backpropagation to each one of them (Wang et al., 2017). This algorithm, introduced in 1990, is called Backpropagation through Time (BPTT) (Wang et al., 2017; Lipton et al., 2015). All the current RNNs use this algorithm (Lipton et al., 2015).

Unfolding is an easy and useful way to view the RNN without cycles (Graves, 2014) (Fig 7).

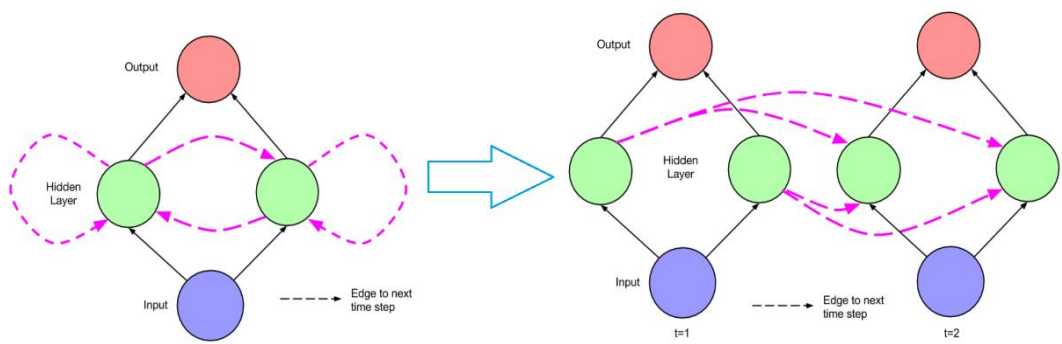

***Figure 7*** *Unfolding Process (Lipton et al., 2015)*

Goodfellow et al. (2016) reported that we can present the process of unfolding by ( g ) function in the following equation:-

$$\begin{aligned}\boldsymbol{h}^{(t)} &= g^{(t)}\left(\boldsymbol{x}^{(t)}, \boldsymbol{x}^{(t-1)}, \boldsymbol{x}^{(t-2)}, \ldots, \boldsymbol{x}^{(2)}, \boldsymbol{x}^{(1)}\right) \\ &= f(\boldsymbol{h}^{(t-1)}, \boldsymbol{x}^{(t)}; \boldsymbol{\theta}).\end{aligned}$$

Where ( g ) function takes all the previous inputs and computes the current state. But we can factorize this function to repeated function ( f ). Goodfellow et al. (2016) states that the unfolding process brings two main advantages:-

1- It can be used the same RNN input size through the timesteps whatever the sequence length is, because, as it can be seen from the previous equation, ( f ) function is a repeated function from one state to another one through the time with the same input size.

2- It can be used the same ( f ) function for each time step.

So, we have just one model to learn each time step, rather than having one model ( g ) for all the time steps.

### 2.3.3.4 Modern RNN architectures

There are two main RNN networks:- Bidirectional Recurrent Neural Networks **BRNN** and Long Short-Term Memory LSTM.

**Bidirectional Recurrent Neural Networks BRNN**

This network was introduced in 1997 (Lipton et al., 2015). BRNN consist of two hidden layers. Both of them are connected to the input and to the output but one of them follows the forward pass and the second one follows the backward pass to form what is called sequence to sequence (s2s) pass (Goodfellow et al., 2016; Wang et al., 2017) (Fig 8).

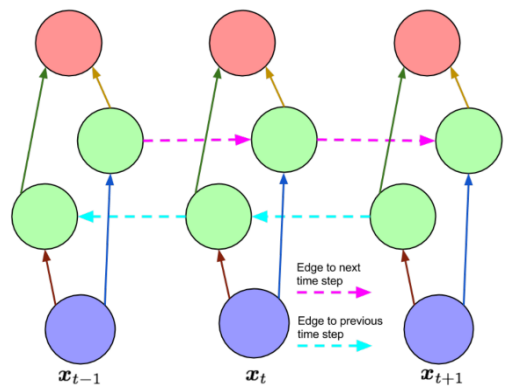

*Figure 8* *Bidirectional Recurrent Neural Networks BRNN (Lipton et al., 2015)*

BRNN have one limitation is that cannot run continuously since it needs two fixed endpoints (one in the past and one in the future)(Lipton et al., 2015). BRNN is widely used in several applications like speech recognition (Goodfellow et al., 2016).

**Long Short-Term Memory LSTM**

LSTM was introduced in the same year as BRNN (Wang et al., 2017). LSTM consists of several components including- input data, hidden state, input state, internal state, the input gate, forget gate and output gate (Figure 9).

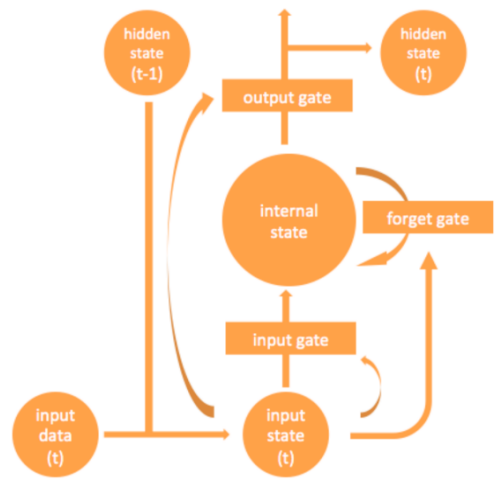

*Figure 9* One LSTM cell (Wang et al., 2017)

- Input data:- the network data denoted as X.
- Hidden state:- the values of a previous hidden layer of the network; it is denoted as h.
- Input state:- the values of the linear combination of the input data in the current time step and value of hidden layer from the previous time step. We can formalize this component as:-
$$i^t = \sigma(W_{ix}x^t + W_{ih}h^{t-1})$$

- Internal state:- it is represented as memory, labelled as m.
  The gates, in general, make the decision for the information of the states.

- Input gate:- makes a decision whether the input state enters the internal state or not. It formalized as
$$g^t = \sigma(W_{gi}i^t)$$

- Forget gate:- makes a decision whether the internal state forgets the previous internal state or not, we have
$$f^t = \sigma(W_{fi} i^t)$$
- output gate:- makes a decision whether the internal state passes its value to the output and hidden state of the next time step or not.
$$o^t = \sigma(W_{oi} i^t)$$

Based on the previous description we can formalize the LSTM as these two equations:-

$$m^t = g^t \odot i^t + f^t m^{t-1}$$

$$h^t = o^t \odot m^t$$

Where $\odot$ is the pointwise multiplication (Lipton et al., 2015). Sometimes called element-wise product (Wang et al., 2017).

This network was invented as a solution to the vanishing gradient problem. Where LSTM's gates are tuned during the learning process to control the follow of information through the network, for example, Forget gate learns to either allow change, not change or partially change the internal state (Jozefowicz et al., 2015).

**2.3.4 Problems with CNN and RNN training**

There are two main problems in DL training by backpropagation:- vanishing and exploding gradients (Nielsen, 2017; Coursera).

- Vanishing gradient happens when the weights are too small and we backpropagate through several layers, then the gradient becomes smaller and smaller.
- Exploding gradient happens when weights are big and we backpropagate through several layers, then the gradient becomes larger.

As this thesis mentioned in section 2.3.3.4, LSTM was invented as a solution to the vanishing gradient problem. In addition, we mentioned in section 2.3.2.2 that ReLU is a solution to this problem as well. Wang et al. (2017) stated that these solutions (LSTM and ReLU) bypass the vanishing and exploding problems by using a clever design, rather than by solving it fundamentally.

Based upon the review of the literature, this project will use the deep learning topologies, CNN and RNN-LSTM, as classifiers to learn how to differentiate patients and healthy subjects.

## 2.4 Related work

Several researchers have worked on the diagnosis of PD by using computational methods, e.g. diagnosis using voice (Das, 2010); diagnosis using brain scan images (Choi et al., 2017); computational analysis of drawings, such as clock drawings (Souillard-Mandar et al., 2016), meander patterns (Pereira et al., 2016; Pereira et al., 2017), spirals (Pereira et al., 2015; Pereira et al., 2016; Pereira et al., 2017 ), and spiral pentagons (Vallejo et al., 2016). The researches ranged between either analysing the PD exam images or study the subject's motor signals by using several approaches like traditional machine learning algorithms (NN, Support Vector Machine (SVM), decision tree and several others) and deep learning (CNN, RNN ).

This section will present some of these works which are related to the main objective of this project.

### 2.4.1 PD with traditional machine learning algorithms

Pereira et al. (2015) employed non-deep learning algorithms in diagnosing PD. Their research aimed to 1-construct a public dataset called "HandPD", 2-create a pipeline for image processing to extract some features from the images, 3- present a new quantitative metric called "Mean Relative Tremor" to measure the tremor of an individual. They emphasised two main tasks:- image processing and feature extraction. The image processing task contains a pipeline of several stages to extract the spiral template ST and the handwritten trace HT, as the image included the template which the subject was asked to follow. Feature extraction aimed to evaluate the differences between the ST and HT by computing nine statistical values between them. They used three classifier models in their experiments:- Naïve Bayes (NB), Optimum-Path Forest (OPF), and SVM.

They constructed a dataset called "HandPD" from 55 individuals: 37 (25 males and 12 females) patients and 18( 6 males and 12 females) healthy controls. This dataset contains handwritten test images for spiral and meander, but in their research, they focused only on the spiral one. In all, they used 373 samples, since they have about four spirals per individual per exam. They divided these samples to 90% training and 10% as a testing set.

They evaluated their works based on recall, precision, F-score and classification accuracy. The results showed that NB obtained the highest accuracy (78.9%) followed by OPF (77.1%) and SVM (75.8%). After completing their work, they noticed that the classifiers did not obtain a good performance in distinguishing the control group compared with the patients group except in NB in recall rate. From our point of view, maybe the weakness in this research is using the traditional machine learning algorithms as the next researches show that using deep learning topologies is obtained better results.

**2.4.2 PD with CNN using time series dataset**

Pereira et al. (2016) explored the use of CNNs to analyse the motor features of PD patients. In other words, they used the CNN on the images extracted from time series signals. Three different CNN architecture were used:- ImageNet, CIFAR-10 and LeNet, as well as OPF as a baseline approach.

They extended the original dataset "HandPD" with signals extracted from a smartpen. They build this new dataset by extracting signals from meander and spiral drawings. This dataset contains 35 individuals, 14 patients (10 males and 4 females) and 21 controls ( 11 males and 10 females). The individuals were asked to fill a form by using a smartpen in order to take the signals from these activities. Then, they transformed the movement signals into images.

They analysed the experimental results using the Wilcoxon signed-rank test and evaluated their work over two image resolutions (64x64 and 128x128 pixels). They used different training-testing set divisions, 75% training and 25% testing, after that 50% for training and 50% for testing sets. In the meander dataset, the imageNet approach obtained the best results (overall accuracy and accuracy per class) in all experiments with both resolutions: ranging between 84% to 87% followed by OPF, ranging between 76% to 84% in overall accuracy. The worst accuracy and recognition (accuracy per class) were achieved by LeNet since it was the shallower net in this experiment. In the spiral dataset, the best accuracies were obtained by OPF, ranging from 77% to 83%, followed by ImageNet, ranging 77% to 80%, and like the meander dataset, the worse results were from LeNet. They noticed that CIFAR-10 obtained 98.53% accuracy in distinguishing PD patients in 50% training-50% testing division and 128x128 resolution.

After completing their research, they noticed that the recognition (accuracy per class) results between PD patient and control were better in the spiral dataset than in the meander one, since the spiral is more difficult to draw than a straight line. They noted that sometimes 64x64 image resolution was not enough for the whole time series data to be distinguished by CNN. From our point of view, the dataset in this research was imbalanced since they have more healthy samples than PD patients so considerable number of healthy people were classified as patients. And using classification accuracy as an evaluation metric, in this case, limits their results analysis, this thesis will discuss accuracy limitation in section 4.4.3.

### 2.4.3 PD with CNN using imaging dataset

Pereira et al. (2017) investigated the use of CNNs to aid PD diagnosis using meanders and spiral images. They compared the application of CNN in several ways including:- standard CNN as a baseline (without optimization); CNN with Bat Algorithm BA (with optimisation); CNN with Firefly Algorithm (FA) (with optimisation); CNN with Particle Swarm Optimisation (PSO) and CNN with random search (RA).

They used the extended dataset "HandPD". They divided this dataset into two groups, spiral and meander groups. The image resolution is 256x256 pixels. Each dataset was divided into 30% as a training set, 20% as a validation set and 50% as a testing set.

Based on Wilcoxon signed-rank test, their results for meander dataset showed that CNN with BA, PSO and standard CNN obtained the best results, (79.62%, 75.76 %, 78.18% ) respectively, among the other techniques as overall accuracy. BA obtained the best accuracy in classifying the PD class. Similarly, the same previous set obtained the highest accuracies, (87.20%, 88.33%, 89.55%) respectively for the spiral dataset. Like the previous experiment, the BA obtained the best accuracy in classifying the PD class.

As in the previous work, they noticed that spiral test is more difficult for patients than meander test, since this test, meandering one, is just to follow a straight line. That is why the experiments on spiral obtained more significant recognition of the healthy from the patients than the meander test.

### 3.4.4 PD with RNN

Arvind et al. (2010) focused on analysing resting tremor to detect PD patients. They extracted the power spectral density (PSD) from electromyography (EMG) by using two methods, PSD by Welch and PSD by Burgs. EMG is a measure of the electrical activity produced by muscles. They used recurrent Elman neural network (REN) model to evaluate the statistical importance of PSD features.

They obtained this EMG signals from the Neurology Department, Sri Ramachandra University, Chennai, India. The EMG recordings were 30 min duration with 100 Hz sampling frequency and obtained from people between 20-30 years. They used 2000 EMG samples as a training set and 1200 as a testing set.

They made the experiments using maximum and mean PSD, calculated by Burgs and Welch methods with a different order of predictor values. The results showed that classifiers obtained 95.66% for Burgs method (PSD mean) with predictor order 5 and 90.41% for the Welch method (PSD mean) with the same predictor order. A potential limitation of this study is that they only used recordings from young people, rather than older people, who are far more likely to suffer from PD.

## 2.5 PD diagnosis problems

The review of the literature indicates that there are some common problems and challenges facing researchers in PD diagnosis, including:-

1- It is very difficult to discover the symptoms in the early stage of the PD, which makes the diagnosis from the hand-drawn images are not enough, as Pereira et al. (2015) found that the hand-drawn exams of both healthy and early-stage patients are similar. This point enlarged especially in tests with meander drawing since it is simpler than spiral one as Pereira et al. (2016) and Pereira et al. (2017) noticed.

2- Datasets are small, biased and rare. This is with the previous point is the reasons for the difficulties involved in the diagnosis from just hand-drawn images. Smartpen can be used to extract more features such as motor features to help us in the diagnosis of PD as Pereira et al. (2016) did in his research.

3- It is difficult to monitor the patient in the home because at-home tools are expensive and not available all the time. This is what Lones et al. (2017) have solved by developing a monitoring device to measure the dyskinesia to help the clinicians in medication regimens.

# Chapter 3. The datasets

This chapter describes the datasets used in this research project.

## 3.1 Time Series Datasets

These datasets were provided by my supervisor, Dr Michael Lones. The data was collected by clinicians at Leeds NHS Trust. All the subjects were asked to draw a copy of a cube drawing and a copy of a spiral pentagon using a smart pen and a tablet (20.3 cm x 32.5 cm) recording the movements during the drawing process. The table 3.1 below shows the datasets' attributes with some rows as an example.

| timestamp | x coordinate | y coordinate | pen angle in x plane | pen angle in y plane | pressure |
|---|---|---|---|---|---|
| 0 | 0.600615155255 | 0.579739793141 | 0.67510199546 | -0.49513813853 | 0.052785925567 |
| 8 | 0.600688906069 | 0.579739793141 | 0.67510199546 | -0.49513813853 | 0.075268819928 |
| 9 | 0.600787410029 | 0.579739793141 | 0.66925722360 | -0.50582969188 | 0.092864125967 |
| 24 | 0.600910441080 | 0.579739793141 | 0.66925722360 | -0.50582969188 | 0.120234601199 |

*Table 3. 1 Time-series dataset example*

It can be noticed that, from table 3.1, coordinates and pressure values are between [0,1], pen angles in the range [-1, 1] and timestamp is an integer starting from 0. We can interpret from zero pressure that the pen was not in contact with the paper; we will discuss zero-pressure in section 5.2.

Our datasets are imbalanced since each dataset includes 58 patients and 29 control subjects. In addition, in the spiral pentagon dataset, each subject had four chances to draw the shape (Fig 10a) using each hand twice. For each patient, the metadata indicates if the patient has PD with normal cognition, PD-MCI (mild cognitive impairment) or PDD (dementia) but this information is not used in this project. Table 3.2 presents the range of time steps in our datasets:

| Datasets | Shortest sample | Longest sample |
|---|---|---|
| Cube | 1268 | 15968 |
| Pentagon | 2662 | 25532 |

*Table 3. 2 Longest and shortest sample in cube and pentagon datasets*

## 3.2 Imaging datasets

We produce imaging datasets from the time series datasets described in section 3.1 using a piece of code in the python script. We produce cube and pentagon images after some data pre-processing, described in section 5.1.2.1, to the original time series datasets. These imaging datasets have the same number of control and patients as the original time series datasets. We will explain the data pre-processing and images drawing process in detail in section 5.1.2.

Fig 10a is the spiral pentagon template that subjects were asked to follow, Fig 10b is an example of a control (LEEDS_c13010514) drawing from the pentagon time series dataset and Fig 10c is an example of a patient (LEEDS_p40310714) drawing from the cube time series dataset.

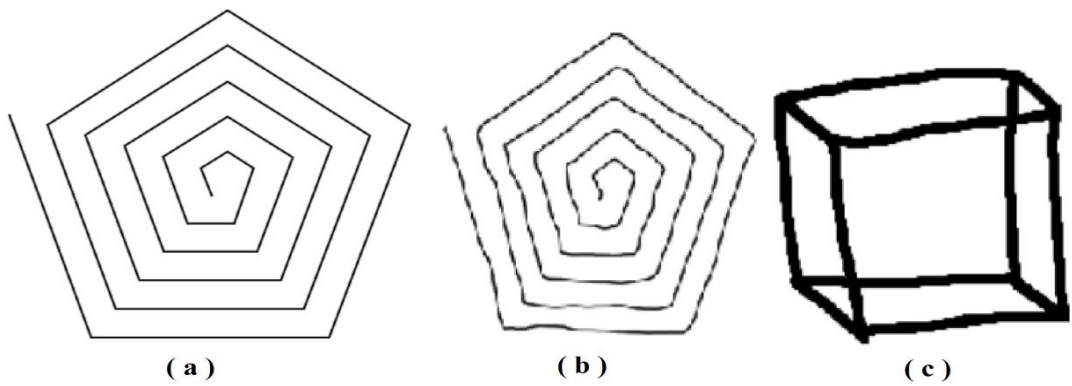

( a )   ( b )   ( c )

***Figure 10*** *the spiral pentagon template (10a), an example of a control (LEEDS_c13010514) drawing (10b) and an example of a patient (LEEDS_p40310714) drawing (10c) from pentagon and cube time series datasets*

# Chapter 4. Methodology

This chapter gives a high-level overview of our workflow, starting with data processing and finishing by evaluating and analysing the results. Details of individual methods can be found in section 5.1.

## 4.1 Data processing

This section presents different necessary types of data and image processing that was applied to the datasets explained in the previous chapter.

### 4.1.1 Data pre-processing

Data pre-processing is an important step in the machine learning process. This step involves tasks like data cleansing which includes checking the NULL values (i.e. empty values) and noise. In addition, using this step we can delete the undesirable features within the dataset. Data pre-processing generally leads to more accurate machine learning results (Kotsiantis et al., 2006).

### 4.1.2 Data preparation (also called data wrangling)

This involves formatting the data correctly to be used as input to machine learning algorithms like CNN and RNN since different algorithms require different input formats (Endel et al., 2015).

### 4.1.3 Image processing

1. **Augmentation**

Augmentation is the process of producing more images from a small set of images in order to increase the number of samples in the dataset. The produced images from an augmentation process are copies of the original images with some modification, such as random rotation, random shift left and right, random zoom with a certain value and random horizontal flip. Augmentation helps avoid overfitting since the model will not see the same sample twice, and it achieves better classification since more samples generally lead to better learning (Ahmad et al., 2017). Fig 11 shows the original (Fig 11-left) and augmented (Fig 11-right) images.

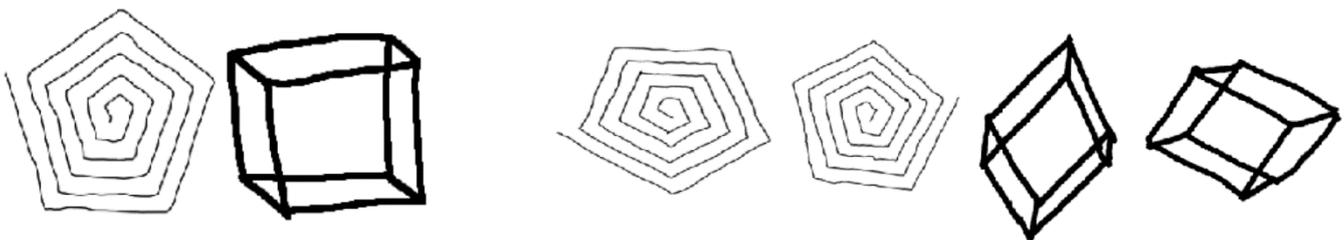

*Figure 11* subject LEEDS_c13010514 and LEEDS_p40310714(left) and some augmented images (right)

2. **Resizing images** this process aims to produce new images with different resolution. The image resolution means the number of pixels within the image, e.g. 32x32 pixel.

3. Image Normalization (standardization)

This technique is used to change the range of image pixel values. This step aims to speed up the learning process as the input transformation of the classifier improves its performance (faster convergence). One common technique is computing the mean and standard deviation (Std) of the images followed by subtracting each image's pixel values from the computed mean and dividing by the computed Std will make the images have zero mean and very small range of pixel values (Wiesler et al., 2011; Ioffe et al., 2015).

In the time series datasets, we apply some data pre-processing including checking, cleaning then data preparation (i.e. preparing RNN input). In the imaging datasets, we apply some data pre-processing including checking, cleaning, then drawing the images. After that, we apply image processing including augmentation, resizing and normalization, then data preparation (i.e. preparing CNN input). We will describe these processes practically in section 5.1.

## 4.2 Signal-based classification

The RNN model will be used on the time-series datasets mentioned in section 3.1. This RNN model will be trained on the hand signals (patients and healthy) extracted from a smart pen during the PD test. These signals contain information about motor features of this subject. The RNN will be trained by using BPTT, which was described in section 2.3.3.3. The model will learn from the controls and patients signals to classify and differentiate between PD class and healthy class for un-seen (new) individuals to determine if this new subject has PD or is healthy.

The rationale behind choosing RNNs is that our time-series datasets have very long and different sample lengths which makes these datasets difficult to mine effectively. And RNNs-LSTM can learn what to remember from such sequences. That is why RNN, in particular, LSTM hidden units are powerful and widely used for learning from sequence data (Lipton et al., 2015).

## 4.3 Image-based classification

In this classification, the CNN model will be used on the hand-drawn images from the datasets mentioned in section 3.2. The CNN model will be trained on these drawings (patients and healthy) by using backpropagation and will then be used as a classifier to differentiate between healthy subjects and patients.

## 4.4 Evaluation

The evaluation process provides a deep understanding of the performance of the methods which were developed in the project. Our evaluation plan consists of two phases: evaluating the experiments' results, then evaluating the models in the experiment with the best results. Before starting the evaluation phases, we should describe cross-validation which is the key technique that enables us to evaluate the classifiers accurately.

### 4.4.1 K-fold Cross-validation (CV):

In this technique, the dataset is divided into K folds (e.g. K=10 in CV-10) and the model is trained on all these folds except one, which is used as a validation set. We repeat this process for all the folds by taking a different fold each time as the validation set (Brownlee, 2013). This technique enables us to collect K different results from each experiment (in our case K=10). For each experiment, we collect many metrics (as we will see in the following sections), but we use Kappa as a basic metric. We will define Kappa and discuss this choice in section 4.4.3.

### 4.4.2 Evaluating the Experiments

Using 10-fold cross-validation enables us to provide a **statistical analysis** of the results as it gives us 10 different results (Kappa values) from each experiment. We used three statistical analysis tests to compare our experiments' results.

1. **Mann-Whitney U Test two-tailed**: is a non-parametric test which helps to compare the distribution of two sets of values using a probability value p-value. This test is used when the two compared groups are not normally distributed and we have two independent groups. In this test, we assume two hypotheses, null and an alternative hypothesis.

    The null- hypothesis **H0**: there is no significance between the two groups.

    Alternative hypothesis **H1**: there is a significance between the two groups.

    p-value: 'probability of incorrectly rejecting a true null hypothesis'. In other words, it measures the compatibility of the statistical model (two compared groups) with the null hypothesis. A P-value <= 0.05 indicates strong evidence against the null hypothesis (we can reject the null hypothesis) and choose the group with the higher median. A p-value > 0.05 means we do not have evidence to reject the null hypothesis and we can choose the suitable group visually using

violin plot which is described in the following subsections, we will discuss these situations in section 6.1.

Mann-Whitney is an appropriate test in our case since we split the dataset every experiment into training & validation and testing sets randomly. So, for each experiment, our model will train on a different set and test on a different set. Therefore, we have independent experiments. In addition, our results in each experiment are not normally distributed (Corder et al., 2014).

2. **Kruskal-Wallis Test:** this test is an expansion of the Mann-Whitney test. In this test, we can compare more than two independent groups. It is identical to Mann-Whitney test in terms of H0 & H1 hypothesis and p-value, except that p-value < =0.05 indicates that there is significance between the three groups, but this test could not specify which one is the dominated group. Therefore, we also need to perform a post hoc test (Corder et al., 2014).

3. **Post hoc test:** post hoc means 'after the fact'. There are many post hoc tests but we chose the Tukey test honestly significant difference (HSD). Tukey test is well known and widely used. This test gives us which group is the significant group (NIST, 2013). Section 6.1 will discuss using the Tukey test.

In order to evaluate the experimental results visually, we use the following 3 plots and descriptive statistics, as depicted in Fig 12.

1. **Statistical summary schedule:** it consists of "five number summary". These numbers, as Fig12 shows, are Minimum (lower whisker), 1st quartile (25% of a dataset), median (2nd quartile, 50% of a dataset), 3rd quartile (75% of a dataset) and maximum (upper whisker).

2. **Notched box plot:** it is a variant of the boxplot. It is a useful plot since, according to (Chambers, 2017), two groups' medians differ (with 95% confidence) if their two notches are not overlapped. The notch width represents the confidence interval (Chambers et al., 2017).

3. **Violin Plot:** it is considered as a generalization of boxplot, where it shows the dataset distribution with a curve line and we can notice the median is depicted as a small white circle and the black box is the difference between $1^{st}$ quartile and $3^{rd}$ quartile (as called interquartile) (Hintze et al., 1998).

4. **Bean plot:** it shows the smoothed distribution density and the individual measurements in the dataset (the small white horizontal lines) (Kampstra, 2008). Each one of these plots, as shown in Fig 12, highlights different aspects of our results. Therefore, we decided to use all of them in evaluation and analysis of our results.

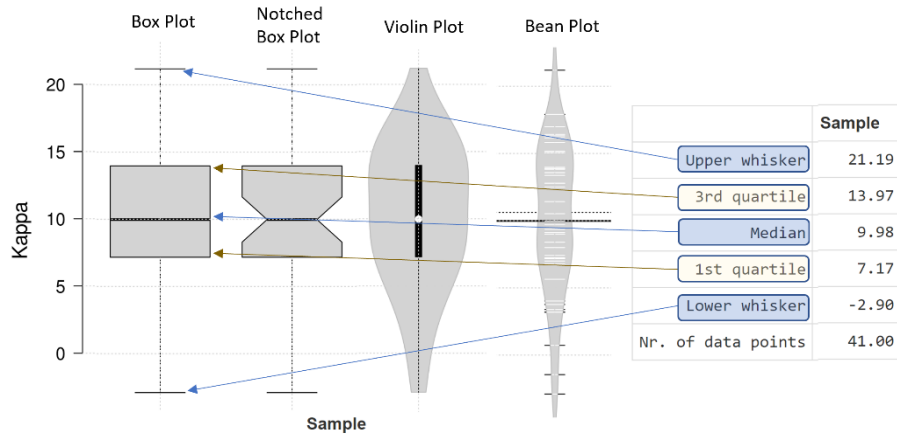

*Figure 12* Box plot, notched box plot, violin, bean plots and statistical summary example

### 4.4.3 Evaluating the Models

Once we choose the best deep learning technique and dataset from the previous phase, we use several metrics to evaluate the models produced by the 10-fold cross validation within the winning experiment. Before describing the metrics, it is essential to know about the confusion matrix.

1. **Confusion matrix** A confusion matrix is a table used usually in classification problems. This table (Fig 13 ) represents the instances in an actual class as rows and the instances in a predicted class as a column. There are four quantities in this table: the number of True positive (TP), True negative (TN), False positive(FP) and false negative (FN).

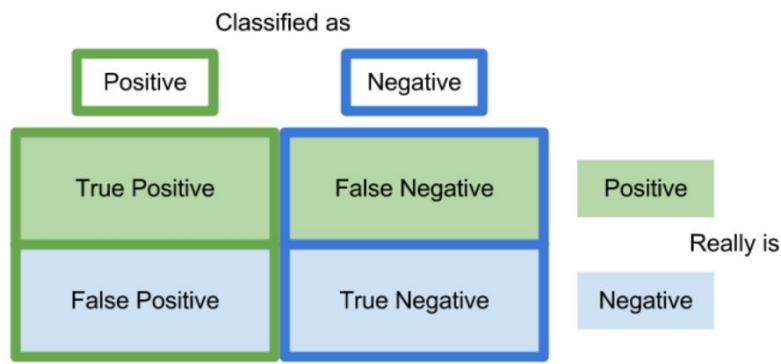

*Figure 13* Confusion Matrix (Greenfield, 2012)

- True positive (TP):- the number of instances which belong to the target class and are correctly classified as the target class.

- True negative (TN):- the number of instances which do not belong to the target class and are correctly classified as not the target class.

- False positive(FP):- the number of instances which do not belong to the target class and are misclassified as the target class.

- False negative (FN):- the number of instances which belong to the target class and are misclassified as not the target class.

Now we can analyse the confusion matrix by calculating several metrics including:-

2. **Specificity** (SP) or true negative rate (TNR):- (Fig 14)

$$SP = TN / (TN + FP)$$

3. **Sensitivity** (SE) or true positive rate (TPR) sometimes called recall:- (Fig 14 and 15)

$$SE = TP/(TP+FN)$$

4. **Precision** (P):- (Fig 15)

$$P = TP/(TP+FP)$$

5. **Classification accuracy** (CA):-

$$CA=(TP+FN)/\text{total number of instances}$$

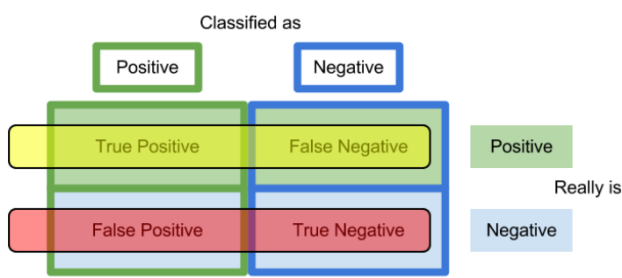

*Figure 15* Sensitivity and Specificity (Greenfield, 2012)

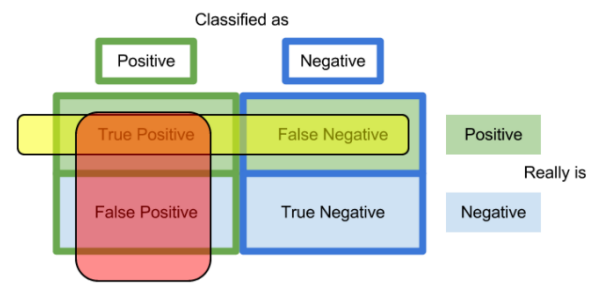

*Figure 14* Precision and Recall (Greenfield, 2012)

6. **Kappa** is a statistical measurement of the agreement between two rankers. In our case (i.e. machine learning) the rankers are the original class (ground truth) and the predicted classification using the classifier (Fig 13) (Landis et al., 1977). There is no standard way to interpret Kappa value, but Fleiss (1981) considered that Kappa > 0.75 is excellent, 0.40-0.75 is fair to good, and < 0.40 as poor agreement. Kappa value could be negative when there is no effective agreement between the two rankers (Fleiss et al., 1981).

We calculate Kappa using the following equations:

$$\text{Kappa} = (\text{observed accuracy} - \text{expected accuracy})/(1 - \text{expected accuracy})$$

$$\text{Total} = TP + FP + TN + FN$$

$$\text{Observed accuracy} = (TP + TN) / \text{Total}$$

$$\text{Expected accuracy} = (TN + FP)*(TN + FN) + (FN + TP)*(FP + TP)/ \text{Total}*\text{Total}$$

The reason behind using Kappa as a basic metric instead of classification accuracy is that our datasets are imbalanced, so normally the testing set consists of an imbalanced number of controls and patients. This means that classification accuracies can be misleading. For example, consider a test set of 3 controls and 6 patients and the classifier predict all of them as patients. In this case, the classification accuracy is about 66% while kappa is 0%. It can be seen that kappa gives us a strong indication that this classifier is not good. To see Kappa values of our experiments, please see section 6.1 or Appendix I.

**Precision-Recall curve:** this curve shows the trade-off between precision and recall, where high precision refers to low false positive rate FP (i.e. accurate results) and high recall refers to low false negative rate FN (i.e. more positive results). Therefore, the large area under this curve means that the classifier has a high performance; due to the high values of both precision and recall (Pedregosa et al., 2011).

7. **Average precision score AP**: sometimes called a precision-recall score and it summarises the Precision-Recall curve. This score corresponds to the area under the Precision-Recall curve.

$$AP = \sum_{n}(R_n - R_{n-1})P_n$$

Where R and P are the recall, precision respectively at nth threshold (Pedregosa et al., 2011).

Both the precision-recall curve and average precision scores are widely used as useful measures of prediction success when the classes are very imbalanced, which is true in our case.

Fig 16 shows two different precision-recall curves with the AP value of each of them. We plot this curve for each fold in each experiment. For more curves examples please see Appendix I.

8. **Training time:** this is the number of minutes the model takes until finishing the training. We calculated this metric for each fold in each experiment.

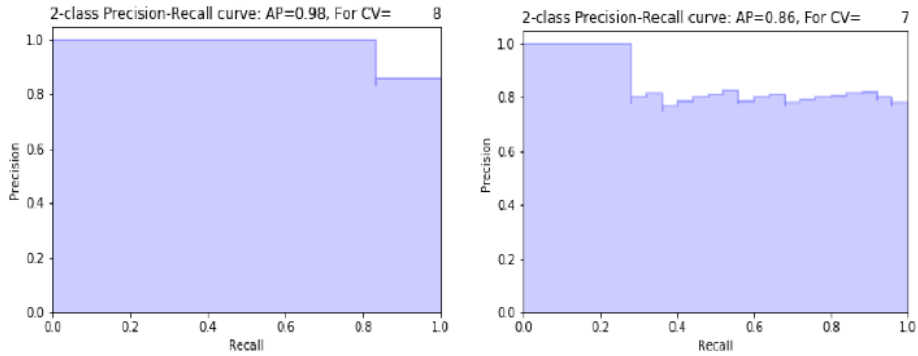

*Figure 16* *Presion_Recall curves with AP values examples*

9. **Plot training accuracy & error VS epoch:** firstly, one epoch represents passing the dataset forward and backward through the neural network once. This plot tells us how the model performance changes, in terms of training accuracy and error, during the epochs. Normally the classification accuracy during the training process would increase with the epochs, while the error would reduce. This means the model is learning. From figure 17-left, it can be seen how the classification accuracy increases (blue line) increase while the error (red line) decreases throughout the epochs and we can notice the test accuracy (horizontal green line). Fig 17-left is one fold of an experiment using CNN on pentagon dataset.

10. **Plot validation accuracy & error VS epoch:** this plot is very similar to the previous plot but the difference is that we draw here the classification accuracy and the error during the validation process instead of training. Fig 17-right is one fold of an experiment using RNN on pentagon dataset.

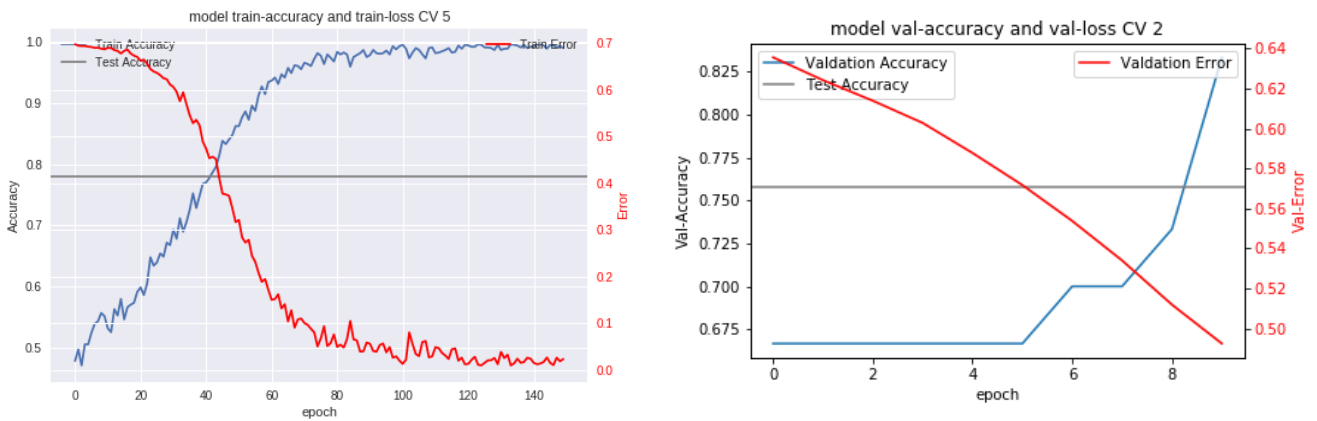

*Figure 17* *Model training accuracy & error VS epochs (left) and Model validation accuracy & error VS epochs*

We draw these plots for each fold in each experiment. These plots help us to determine properly how many epochs we need to train our model since more epochs means more learning (until reaching overfitting) and fewer epochs means less learning (until reaching underfitting).

### 4.4.4 Evaluation plan

Our approach uses a Mann-Whitney U test to compare pairs (like CNN pentagon_32x32 augment VS 32x32 imbalance). The reader can see other pairs from Fig 26. In cases where the Mann-Whitney test tells us that there is significance between the two groups, then the group with higher median can be chosen using the statistical summary schedule as described theoretically in section 4.4.2 and practically in section 6.1. If the test gives no significance then we can choose the suitable group, i.e. the stable and the less risky. We assess this visually using plots like notched box plot, violin and bean plot.

The tripartite comparisons will be performed using the Kruskal-Wallis test. In cases where the Kruskal-Wallis test tells us there is significance between the three groups then we use post hoc test like Tukey to inform us which group is dominated in the tripartite groups. If it gives no significance then again we can choose the suitable group visually.

When we complete the first phase described above by choosing the best experiment and dataset, then the second phase starts by choosing the best model of this experiment since each experiment has 10 models produced by cross-validation. In this phase, we use the metrics described in section 4.4.3. We will discuss this step in more detail in section 6.1.4.

# Chapter 5. Implementation

This section illustrates the implementation of workflows and the experiments.

## 5.1 Implementation Workflow

This section describes our implementation workflows divided into time-series and images classification workflows. These workflows used the methods which are described in chapter 4. Please note that all the following workflows are repeated for the cube and pentagon datasets. Each branch in the workflow is a separate experiment.

### 5.1.1 Time-series Classification Workflow

This workflow, Fig 18, aims to use RNN on our time-series datasets to differentiate between PD class and healthy class as this thesis described in section 4.2. The following subsections will explain the different stages in this workflow starting from the original dataset and finishing by analysing and evaluating the results.

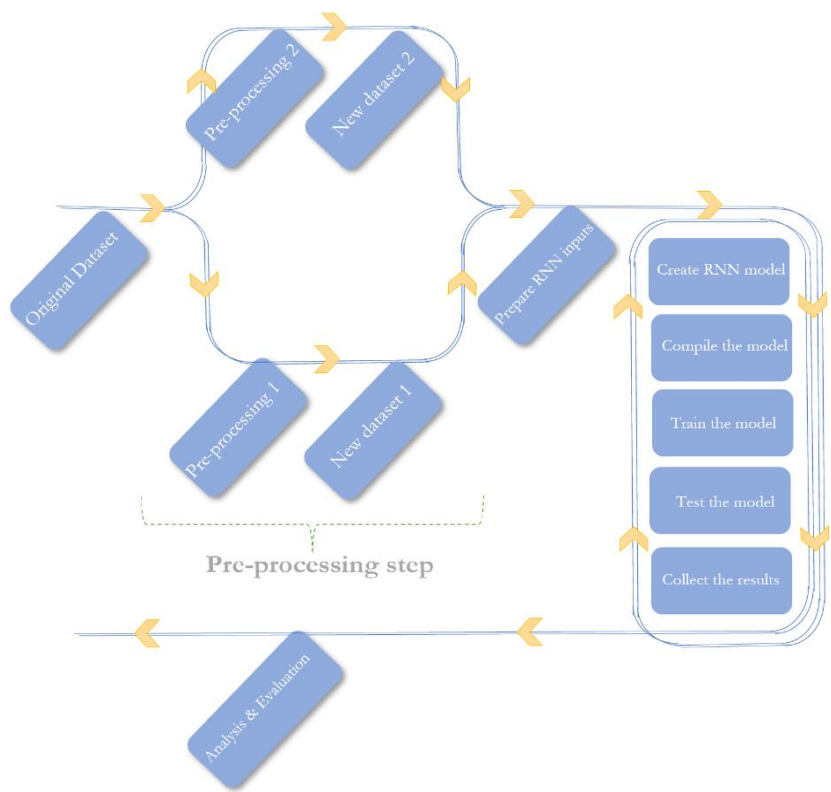

***Figure 18*** *RNN-based experiments workflow*

#### 5.1.1.1 Pre-processing dataset

Our time-series datasets (cube and pentagon) will be processed with the following tasks:

- Either delete the zero pressure (branch 1 in Fig 18) or keep it (branch 2 in Fig 18),
- Delete the empty samples,
- Check if any cell is zero, i.e. any feature at a certain time step.

#### 5.1.1.2 Prepare RNN inputs

This step contains the following tasks:

- Put RNN input in the correct format without taking the timesteps column into consideration (since the sampling rate is known to be constant). RNN input format must be a three-dimensional matrix (#Samples, #Time Steps, #Features) (Brownlee, 2017).
- Assign each sample with its output value (healthy control 0 and patient 1).
- Zero-padding to make all samples the same length (add zeros before each sample).
- Shuffle the inputs with its outputs across the dataset.
- Divide the datasets to 90% training & validation and 10% testing sets.

#### 5.1.1.3 RNN classifier

The tasks in this step are happened for each fold in cross-validation, i.e. 10 times. This step is responsible for the learning, testing and collecting the results processes as follows:

- **Create RNN model:** One LSTM layer includes one LSTM cell with 32 neurons as hidden layer (i.e. internal cell state is a vector with 32 rows) and one dense layer (i.e. fully connected layer) with one unite, sigmoid activation function, as it is binary classification problem (i.e. the model predicts one class).

- **Compile the model:** assign loss function (i.e. objective function, or optimization score function) called binary_crossentropy to the model and choose the stochastic gradient descent (SGD) as optimizer with these configuration: learning rate=0.003, decay=1e-6, momentum=0.9.

- **Train the model:** start the learning process by dividing the dataset for a number of batches, each in which contains 8 samples (batch size=8) and 125 epochs (the learning iterations) for cube dataset and 10 epochs for pentagon dataset. Then we divide the training and validation set to 90% training and 10% validation sets.

- Test the model on the testing set.

- **Collect the results:** we collect these metrics, classification accuracy and error, confusion matrix, precision, sensitivity, specificity, kappa, average precision score, training time, plot training accuracy & error vs epoch, plot validation accuracy & error vs epoch, plot precision-recall curve. All these metrics are described in section 4.4.3.

After we collect the results, then we analyse and evaluate it. Please see section 6.1 to read analyse the results.

### 5.1.1.4 How RNN-LSTM works practically?

In our model, we have an input layer, a hidden layer (LSTM cell), a dense layer and an output layer. For each epoch, we feed the batches one by one to the model. The model will create copies of the network for each sample in the batch to deal with each sample in parallel (the parallelism is only for efficiency). Now for each sample and copy of network, each LSTM cell will be unfolded to LSTM cells with a number equal to timesteps of the sample. Each LSTM cell has internal state (memory) passes through the timesteps forward and this state will be reset when the sample is finished. So, at each timestep, there are 8 states that will be computed for 8 different samples.

When the batch size is completed, the backpropagation (adjustment) will happen. This means that the model updates its weights after training all the samples in the batch size. The model then will take the next batch and feed it to the model and repeat until all the batches are completed. Then the whole process will be repeated for the next epoch until all epochs are completed. (Chollet, 2015).

In our model, we take the output from the last cell (i.e. the last time step). So the 32 neurons in the last LSTM cell are connected to the dense layer. The dense layer gives one output since our problem is binary classification as presented in Fig 19 where n is a number of timesteps.

Figure 19-left shows the 3D view of the LSTM model and its copies. The drawing shows the model in the training phase so in the testing phase we will have only one input, one copy of the LSTM, a dense layer and one output. Each LSTM cell here is a folded cell. While Fig 19-right shows the 2D plan for one copy of the unfolded model through the timesteps (shows only the first and last steps).

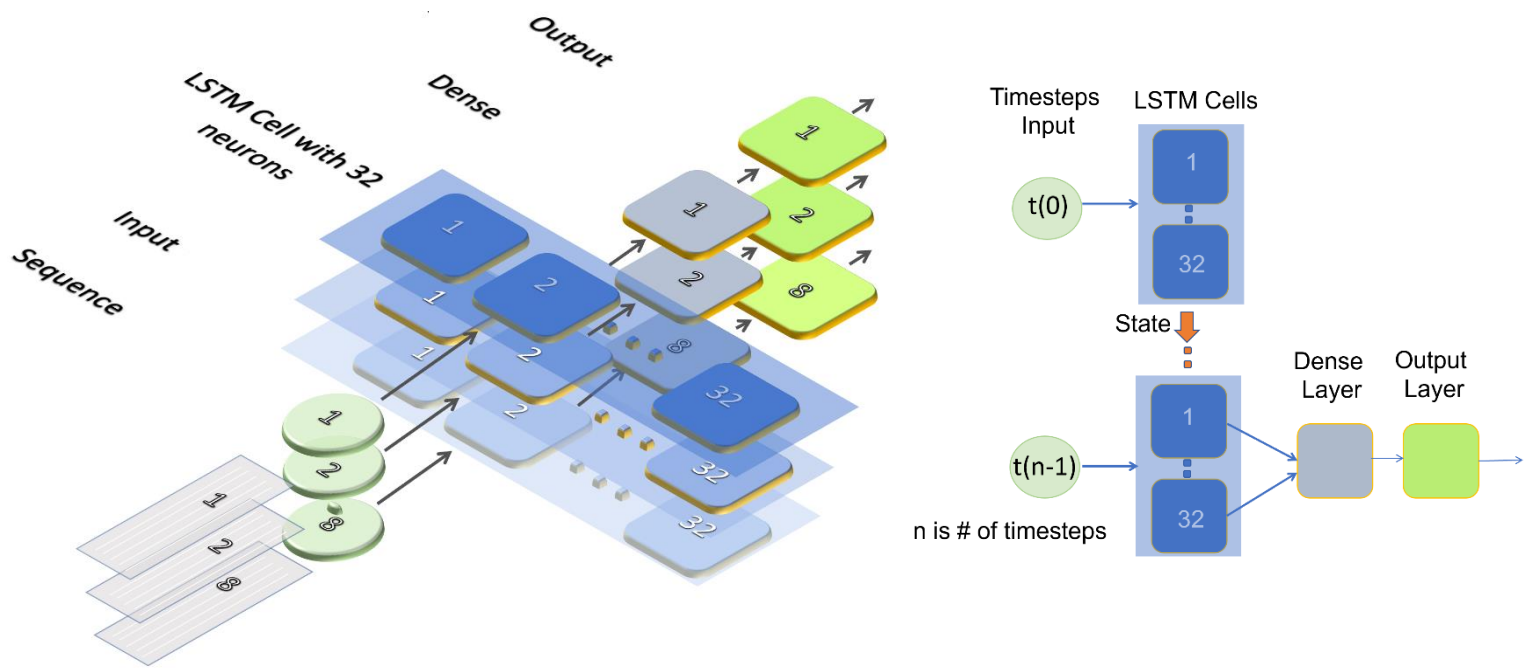

**Figure 19** *The Proposed 3D LSTM Architecture (left) and 2D LSTM Architecture (right)*

### 5.1.2 Images Classification Workflow

This workflow, Fig 20, is designed to achieve the image-based classification described in section 4.3. the following subsections will explain this workflow stages.

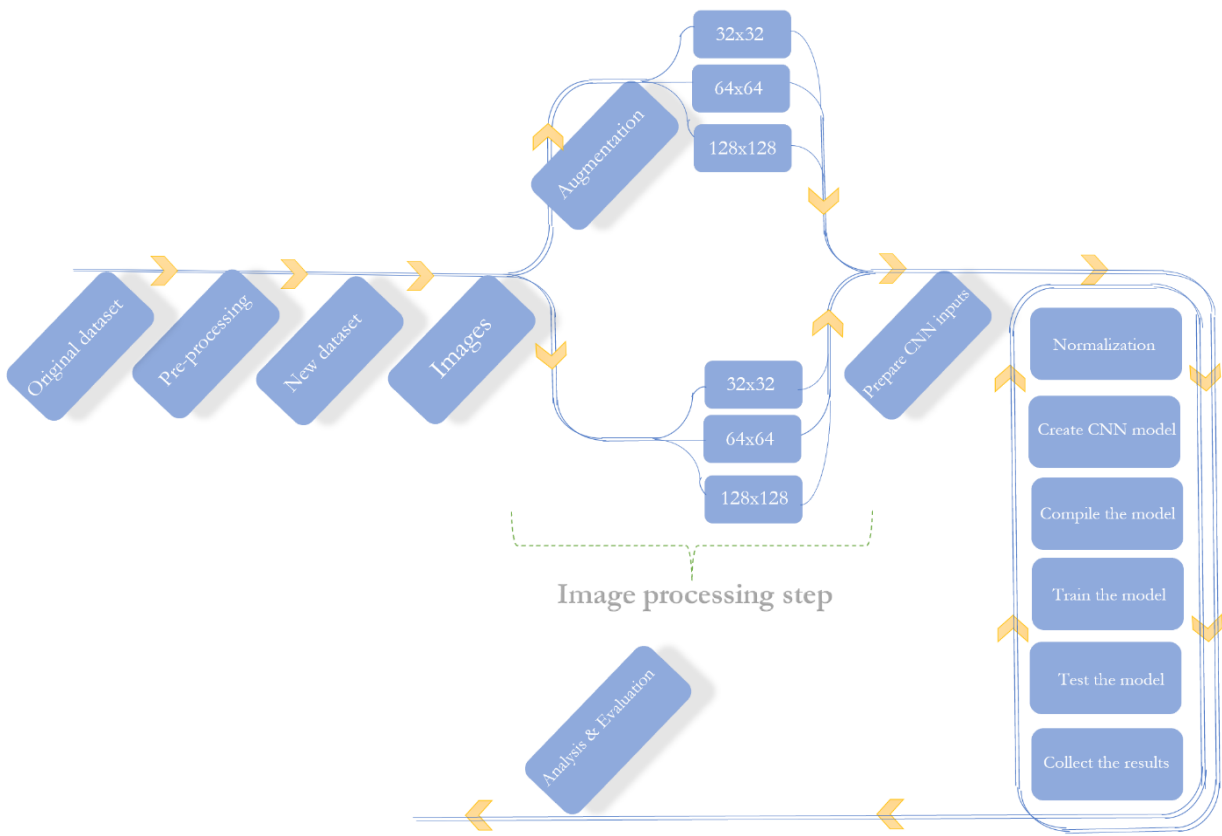

**Figure 20** *CNN-based experiments workflow*

### 5.1.2.1 Pre-processing dataset:

This step is the same as in the RNN workflow in section 5.1.1.1, except the following:

o   We do not need zero-pressure here.

o   Delete the initial waiting in the samples since some subjects wait sometimes before the start of the drawing. So, we make sure there are no extra lines in the images.

### 5.1.2.2 Image processing:

Firstly we draw square grayscale images from the processed dataset with size 288x288 pixel and with pixel values between 0 and 255. At this point, the workflow, Fig 20, branches to either augment the images then resize the produced images or resize the original images directly.

In the upper branch (with augmentation) we produce 23 copies of each control image and 11 copies of each patient image. We applied augmentation with random rotation, random zoom with a certain value and random horizontal flip. The differentiation in the number of copies, i.e. 23 copies VS 11 copies, gives us a balanced dataset.

After the augmentation, the produced images will be resized to three different sizes 32x32, 64x64 and 128x128 pixel. In the other branch, we resize the original images (288x288 pixel) directly to 32x32, 64x64 and 128x128 pixel. So we keep the imbalance in the dataset.

### 5.1.2.3 Prepare CNN inputs

This step contains the following tasks:

o   Put CNN input in the correct format. CNN input format must be a three-dimensional matrix (#channels, width, height ) (Brownlee, 2017). For example, for one of our experiments, the input is (1, 32, 32). Where a number of channels are (1) since we have grayscale images.

o   The rest tasks in this step are as same as in section 5.1.1.2 except we do not need padding here.

### 5.1.2.4 Image Normalization

In this step, we compute the mean and standard deviation (Std) from only the training set. After that, we subtract the mean from each pixel value of images in training, validation and testing sets. and then dividing the results by the standard deviation. So the pixel values lie in the range from negative to positive and the data centre lies at zero as we described in section 4.1.3.

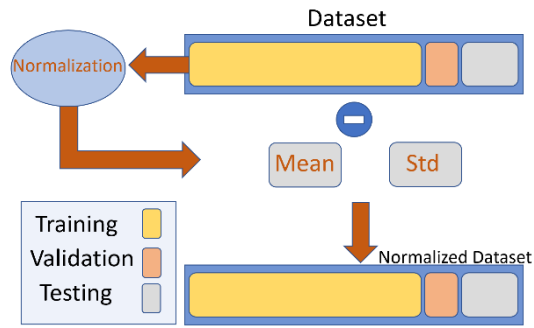

**Figure 21** *Normalization process*

From Fig 21, the mean here is an image. Each pixel of this image represents the mean of all pixels at the same position in the training set. i.e. the first pixel represents the mean of all first pixels of the training images. Std is the same concept but each pixel is Std of all pixels at the same position (Wiesler et al., 2011; Ioffe et al., 2015). In addition, for each fold, we have random splits, i.e. we have different training – validation pairs, This is why we perform a normalization process for each fold. This randomization enables us to collect independent results.

**5.1.2.5 CNN classifier**

As in section 5.1.1.3, The tasks in this step are happened 10 times. This step is responsible for the learning, testing and collecting the results processes as follows:

- **Create CNN model:** our model consists of 6 convolution layers, 3 max-pooling layers, 6 dropout layers, 3 dense layers and one flatten layer. All the activation functions are Relu except the last dense layer with sigmoid activation function. Fig 22 shows our model.

Drop out layer is a technique introduced by Srivastava et al. (2014). This layer aims to avoid overfitting by ignoring randomly some neurons from the previous layer. So, these neurons will not contribute to the learning process. We add this layer after first each of con32, con64, conv128 and after flatten, Dense1024 and Dense512 layers. We chose this particular CNN architecture since it gives good results in (Pereira et al, 2016; Brownlee, 2016).

The rest tasks here are same as in section 5.1.1.3 but with batch size=16 and 150 epochs for both pentagon and cube datasets. Please see section 6.1 to read analyse the results.

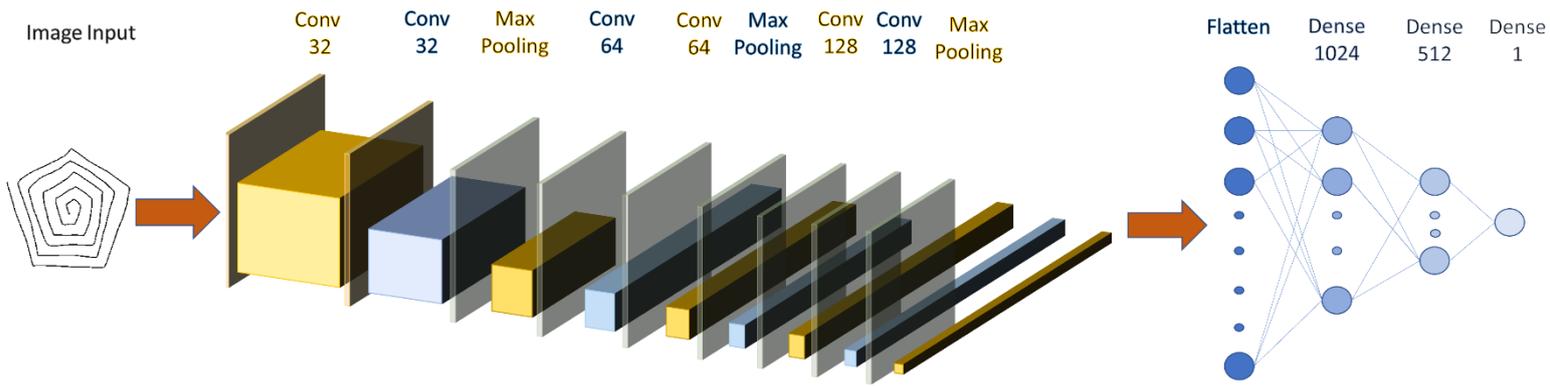

*Figure 22* The Proposed CNN Architecture

**5.1.2.6 How CNN works practically?**

For each epoch, we feed the batches one by one to the model. The model will create copies of the network for each sample in the batch to deal with each image in parallel. When the batch size is completed then the backpropagation will happen at this point. This is means model updated the weights after train all the images in the batch size. The model then will take the next batch and feed it to the model and repeat until complete all the batches. And then the whole process will be repeated for the next epoch until complete all epochs.

Why we choose all these configuration values exactly?

Most of these configuration values are chosen empirically. One can be wondered why the RNN is trained using only 10 epochs on the pentagon dataset. This is because choosing 125 epochs, like in using RNN on cube dataset, will lead to 6 days of training time which is infeasible in this project's period and our current machine. To choose the optimal values that give the highest performance & results we could use a meta-heuristic (optimization algorithms) such as an evolutionary algorithm to choose these values.

## 5.2 Experiments Explanation

In order to achieve the project's objectives mentioned in section 1.2, we completed 16 separate experiments, presented in tables 5.1, 5.2, 5.3 and 5.4. we can group our experiments as follows:

- **RNN Experiments**

These experiments are grouped based on the pressure feature as follows:

- Experiments without zero-pressure consideration
- Experiments with zero-pressure consideration

These experiments aim to compare the subject's motor signals only during the drawing task and throughout all the exam (including the pen is not in contact with the paper). For this, we completed 4 different experiments using the RNN technique on both pentagon and cube datasets as shown in tables 5.1 and 5.2.

- **CNN Experiments**

These experiments are grouped based on the distribution of the datasets as follows:

- Experiments with imbalanced datasets
- Experiments with balanced datasets (i.e. with augmentation)

These experiments aim to investigate the effect of the dataset's distribution on the classifier performance and explore this effect with a variety of image sizes. We completed 12 different experiments with CNN technique on both pentagon and cube datasets, as it can be seen from the tables 5.3 and 5.4.

Tables 5.1, 5.2, 5.3 and 5.4, show the number of samples in training, validating and testing sets; the number of controls and patients and the total number of samples. Tables 5.1 and 5.2 show the number of samples' timesteps in each experiment, and tables 5.3 and 5.4 show the sizes of images which we used in our experiments.

It can be noticed some differences in numbers between the subjects(controls and patients) in the tables below and what we have presented in the datasets description in chapter 3. This is because there are some empty samples in the original datasets and the values in the tables below are taken after processing. In addition, we considered each attempt by a patient in the pentagon dataset as a sample (i.e. one subject).

| RNN_Pentagon | Timestamps | Features | Train | Validate | Test | Total |
|---|---|---|---|---|---|---|
| Without 0 pressure | 24969 | 5 | 263 | 29 | 33 | |
| With 0 pressure | 25532 | 5 | 263 | 29 | 33 | |
| | | | Control | | Patient | 325 |
| | | | 102 | | 223 | |

*Table 5. 1* RNN_Pentagon experiments

| RNN_Cube | Timestamps | Features | Train | Validate | Test | Total |
|---|---|---|---|---|---|---|
| Without 0 pressure | 8266 | 5 | 67 | 6 | 9 | |
| With 0 pressure | 15968 | 5 | 67 | 6 | 9 | |
| | | | Control | | Patient | 82 |
| | | | 26 | | 56 | |

*Table 5. 2* RNN_Cube experiments

| CNN_Pentagon | 32x32 | 64x64 | 128x128 | Train | Validate | Test | Total |
|---|---|---|---|---|---|---|---|
| Imbalanced | ✓ | ✓ | ✓ | 263 | 29 | 33 | |
| | | | | Control | | Patient | 325 |
| | | | | 102 | | 223 | |
| **CNN_Pentagon** | **32x32** | **64x64** | **128x128** | **Train** | **Validate** | **Test** | **Total** |
| Augmented | ✓ | ✓ | ✓ | 3438 | 381 | 425 | |
| | | | | Control | | Patient | 4244 |
| | | | | 2079 | | 2165 | |

*Table 5. 3* CNN_Pentagon experiments

| CNN_ Cube | 32x32 | 64x64 | 128x128 | Train | Validate | Test | Total |
|---|---|---|---|---|---|---|---|
| Imbalanced | ✓ | ✓ | ✓ | 67 | 6 | 9 | |
| | | | | Control | | Patient | 82 |
| | | | | 26 | | 56 | |
| **CNN_ Cube** | **32x32** | **64x64** | **128x128** | **Train** | **Validate** | **Test** | **Total** |
| Augmented | ✓ | ✓ | ✓ | 961 | 106 | 118 | |
| | | | | Control | | Patient | 1185 |
| | | | | 588 | | 597 | |

*Table 5. 4* CNN_ Cube experiments

We used Python script version 3 as the programming language to run our experiments and analyse the results. We chose this because it is widely used and strongly supported. Python includes many libraries specialized in deep learning such as Keras[1]. Keras contains many libraries like NumPy and Pandas that help us process the datasets and Sklearn[2] to extract the results from the models. All the plots are drawn using BoxPlotR[3].

We used NVIDIA Tesla K80 (with GPU ~12 GB RAM) from Google Collaboratory[4] to train the models in CNN experiments and our laptop CPU Intel I7 to train RNN models. We used Tesla in CNN since GPU is much faster to process images while CPU is better to train sequences. Table 5.5 shows the experiments' training time and the resources which are used in the training of the networks. The blue rows are experiments with RNN and orange rows are with CNN.

| # | Dataset | Experiments | Resource | Time |
|---|---------|-------------|----------|------|
| 1 | Pentagon | 32x32-Imbalanced | GPU | 7 min |
| 2 | Pentagon | 64x64-Imbalanced | GPU | 14.8 min |
| 3 | Pentagon | 128x128-Imbalanced | GPU | 40 min |
| 4 | Pentagon | 32x32-Augmented | GPU | 1.3 hr |
| 5 | Pentagon | 64x64-Augmented | GPU | 3 hr |
| 6 | Pentagon | 128x128-Augmented | GPU | 8.4 hr |
| 7 | Pentagon | Without 0 pressure | CPU | 11 hr |
| 8 | Pentagon | With 0 pressure | CPU | 11.4 hr |
| 9 | Cube | 32x32-Imbalanced | GPU | 2 min |
| 10 | Cube | 64x64-Imbalanced | GPU | 4.5 min |
| 11 | Cube | 128x128-Imbalanced | GPU | 11 min |
| 12 | Cube | 32x32-Augmented | GPU | 23 min |
| 13 | Cube | 64x64-Augmented | GPU | 54 min |
| 14 | Cube | 128x128-Augmented | GPU | 2.3 hr |
| 15 | Cube | Without 0 pressure | CPU | 12.9 hr |
| 16 | Cube | With 0 pressure | CPU | 24.5 hr |
| Total | | | | 66 hr |

*Table 5.5* Training time for each experiment

---

[1] https://github.com/fchollet/keras
[2] http://scikit-learn.org/stable/index.html
[3] http://shiny.chemgrid.org/boxplotr/
[4] https://colab.research.google.com/

# Chapter 6. Discussion & Evaluation

This chapter analyses the experimental results, discusses the evaluation process, and highlights the research findings.

## 6.1 Results and Analysis

This section presents and analyses the most important results and findings. Full results from the experiments can be found in Appendix I and II.

### 6.1.1 Affect of augmentation

Using the approach described in section 4.4.4, all CNN pentagon and cube comparing pairs (see Fig 26) such as pentagon_64x64_augment vs 64x64_imbalance give a significance with Mann-Whitney test (p_value < 0.05) except cube_32x32_augment vs 32x32_imbalance, which has a p_value = 0.733 > 0.05. Therefore, we choose the group (experiment) with a higher median in the comparisons with significance. While we choose the suitable group in cube_32x32_augment vs 32x32_imbalance. The statistical summary schedule, Fig 23-right, Shows that cube_32x32_imbalance experiment has a higher median than the augmented one. But cube_32x32_imbalance group has high risk since it includes 0% and 100% kappa values in the same experiment. Therefore, we choose cube_32x32_augment approach because it is more stable. The distribution of these two experiments' results can be seen in the violin plots in Fig 23-left including the medians as a small white circle.

We noticed from the results that datasets with augmentation (i.e. balanced datasets) in all cases lead to better results than the imbalanced datasets.

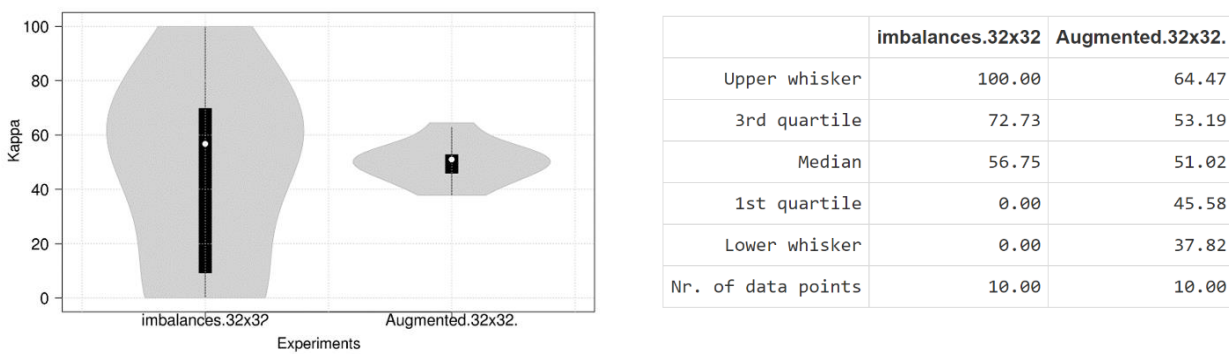

*Figure 23* *Violin plot(left) and statistical summary (right)*

### 6.1.2 Affect of removing zero pressure

There is a big significance between RNN cube with and without zero pressure and it can be noticed easily that the results of the experiment RNN cube with zero pressure were better, suggesting that RNN cube with zero pressure approach is better than RNN cube without zero pressure. Mann-Whitney test found no significant differences in distributions on the RNN pentagon groups so we choose RNN pentagon with zero pressure since it has better results. Cube with zero pressure has better results than pentagon one as there is significance between them and the cube experiment has the higher median.

The results demonstrate that keeping the zero pressure information is meaningful in terms of the differentiation between patients and healthy subjects.

### 6.1.3 Affect of the images size

We use the Kruskal-Wallis test to choose the experiment with the best results from the tripartite comparisons (see Fig 26) like CNN pentagon_32x32, 64x64 and 128x128_augment. Performing Kruskal-Wallis and Tukey tests on this tripartite shows that there is a significance. So, we can reject the null hypothesis at pentagon_64x64_augment as we see in Fig 24-right in the reject column, where Fig 24-right is Tukey test results and Pa64x means pentagon_64x64_augment. Violin plot, Fig 24-left, demonstrates that pentagon_64x64_augment group has the higher median among the other groups. Similarly, we choose 64x64_augment in cube group since it has the higher median between the three experiments.

These results suggest that 64x64 pixel is the suitable size to contain the image data under our CNN configurations.

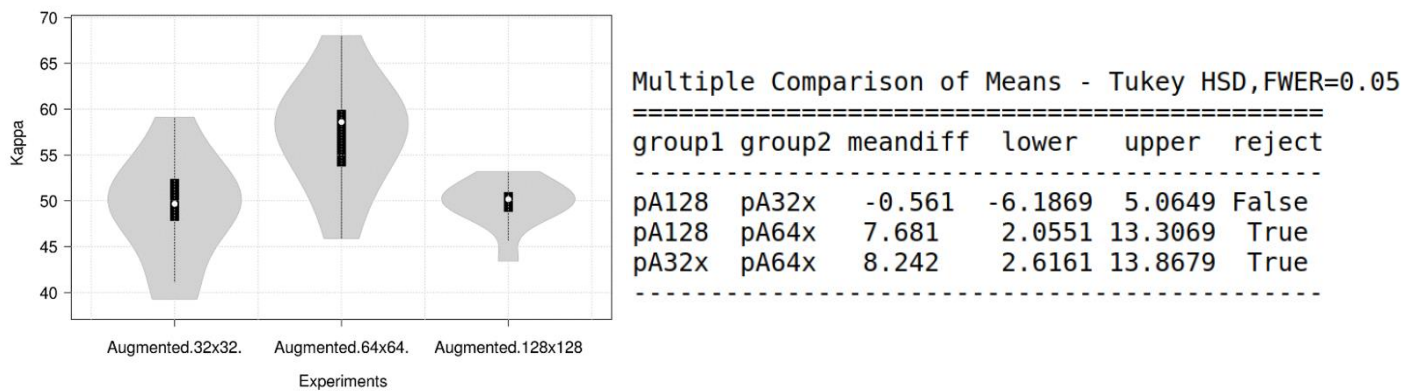

***Figure 24*** *Violin Plot (left) and Tukey test results (right)*

### 6.1.4 Affect of deep learning technique and dataset type

Now, as we have only three experiments at the last level: CNN pentagon_64x64_augment, CNN cube 64x64_augment and RNN cube with zero pressure, the Kruskal-Wallis test shows that there is significance between these three experiments with p-value= 0.00085. But the Tukey test did not help us to determine which experiment is the most significance (Fig 25-top right). Violin plot, notched box plot and statistical summary table enable us to see the comparing matter from many sides (Fig 25).

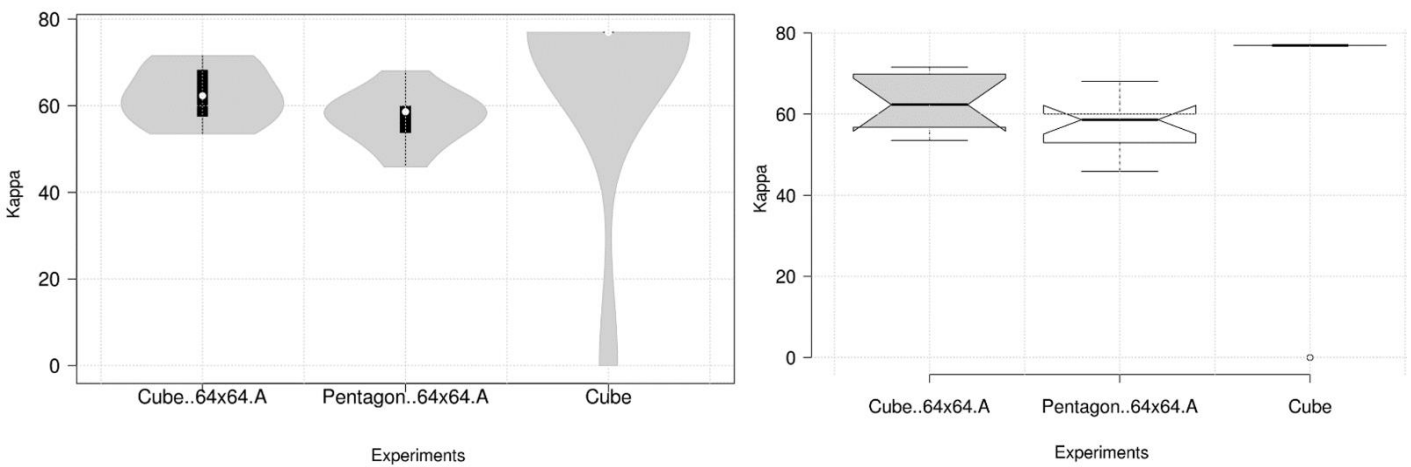

**Figure 25** *Tukey results(top right), statistical summary(top left), violin plot (down left) and notched boxplot (down right)*

Violin plot and statistical summary table in Fig 25 show that cube with zero pressure has the highest median among the other two experiments at 76.91%. The notched box plot shows that all the folds in this experiment have the same kappa value except one outlier has 0% Kappa. This outlier prevents the Tukey test from determining the better experiment since it makes the cube with zero pressure distribution from 0% - 76.91% range of kappa (Fig 25-violin plot), and this leads to include the other two experiments' values in the same range. The other two experiments have good distributions with very close median as it can be seen from the statistical summary table.

Although the followings:

✓ Cube with zero pressure experiment has one fold at 0% kappa as an outlier,

✓ Each approach of the other two experiments, CNN cube_64x64_augment and pentagon_64x64_augment, are good choice to achieve acceptable results in our research to diagnose PD,

✓ Cube with zero pressure experiment takes 24.5 training hours, while the other two experiments have 54 minutes, 3 training hours respectively (table 5.5),

But:

✓ From the previous discussion, it can be seen that cube with zero pressure experiment has 9 folds with 76.91% kappa value,

✓ Cube with zero pressure approach gives much more stable results than the other two experiments,

✓ Cube with zero pressure experiment has higher median which is the most important factor accompanied by the stability,

Thus, under our research conditions including experiments, datasets and training resources. It can be considered that using RNN on the cube with zero pressure dataset gives the best results in our research in diagnosing PD. However, changing at least one of the research conditions might change our decision as our priority would be changed.

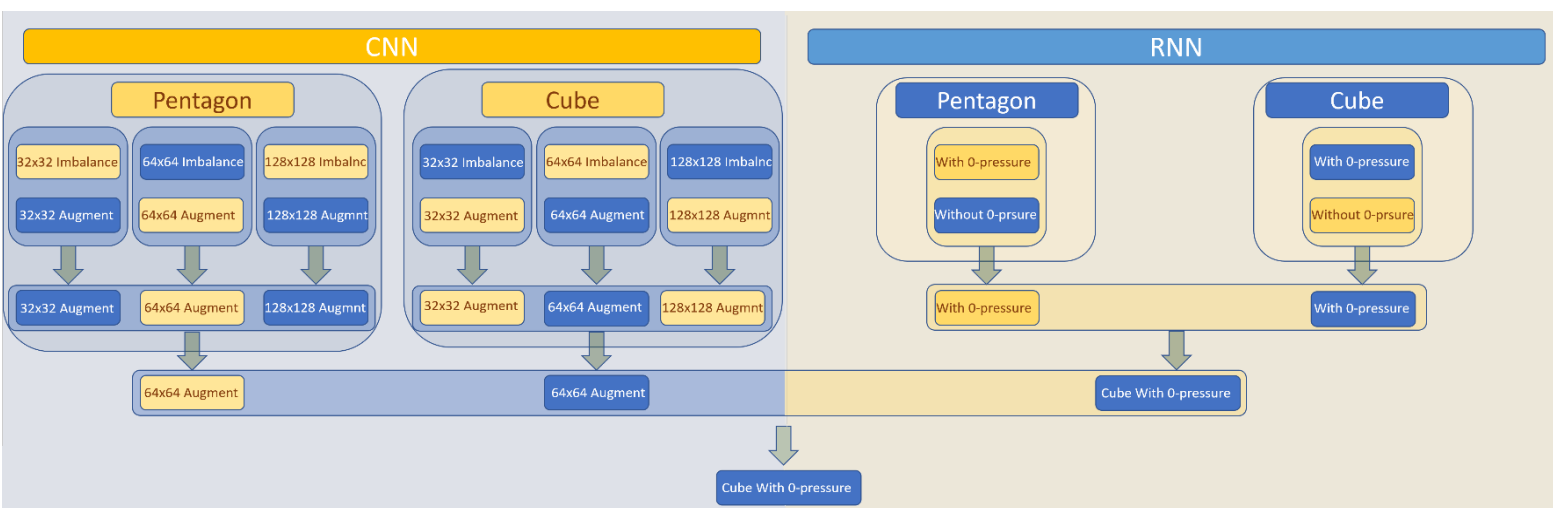

*Figure 26* All experiments comparisons

Now the second evaluation phase starts to choose the model with the better results among the models of RNN cube with zero pressure approach. Table 6.1 shows the metrics in each fold.

Apart from fold 2 (the outlier), it can be noticed easily that all the folds have approximately the same metrics values, but the highest average precision is at fold 8 with 98%. Therefore, it can be considered that this model gives the best performance. Table 6.2 is the differentiation accuracy between healthy and patients in fold 8. And table 6.3 is the confusion matrix of this model. We can see that model 8 successfully classified all the controls and 5 out of 6 patients. The reader can see Appendix I and II for all evaluation plots and results.

| Metric/Fold | 1 | 2 | 3 | 4 | 5 | 6 | 7 | 8 | 9 | 10 |
|---|---|---|---|---|---|---|---|---|---|---|
| accuracy | 88.89 | 66.67 | 88.89 | 88.89 | 88.89 | 88.89 | 88.89 | 88.89 | 88.89 | 88.89 |
| precision | 0.92 | 0.44 | 0.92 | 0.92 | 0.92 | 0.92 | 0.92 | 0.92 | 0.92 | 0.92 |
| recall | 0.89 | 0.67 | 0.89 | 0.89 | 0.89 | 0.89 | 0.89 | 0.89 | 0.89 | 0.89 |
| specificity | 100 | 0.00 | 100 | 100 | 100 | 100 | 100 | 100 | 100 | 100 |
| kappa | 76.92 | 0.00 | 76.92 | 76.92 | 76.92 | 76.92 | 76.92 | 76.92 | 76.92 | 76.92 |
| AP | 0.94 | 0.94 | 0.94 | 0.96 | 0.96 | 0.94 | 0.94 | 0.98 | 0.94 | 0.94 |
| Time(hr) | 2.39 | 2.46 | 2.41 | 2.33 | 2.36 | 2.35 | 2.34 | 2.34 | 2.48 | 2.38 |

**Table 6. 1** *The different metrics in the RNN cube with zero pressure models*

|  | precision | recall | f1-score | support |
|---|---|---|---|---|
| 0 (Control) | 0.75 | 1.00 | 0.86 | 3 |
| 1(Patient) | 1.00 | 0.83 | 0.91 | 6 |
| avg / total | 0.92 | 0.89 | 0.89 | 9 |

**Table 6. 2** *the differentiation accuracy between Patients and Healthy in fold 8*

|  | 0 (Control) | 1(Patient) |
|---|---|---|
| 0 (Control) | TN= 3 | FP= 0 |
| 1(Patient) | FN= 1 | TP= 5 |

**Table 6. 3** *The fold 8 confusion matrix*

### 6.1.5 Discuss the results

Analysing the augmentation results, section 6.1.1, shows that the augmentation process affects positively on the PD diagnosing results. the equal contribution of the two classes in the learning process, perhaps, is the reason behind this good effect. The different image resolutions show approximately similar behaviour in terms of PD diagnosis, although 64x64 pixel was a little better among the other resolutions (section 6.1.3). maybe this size fits suitably with the CNN configurations such as kernel and stride sizes.

It seems the subject's movement signals that have been taken during the drawing process is insufficient to differentiate between patients and healthy subjects. While the movement signals which are obtained during the whole PD exam period affect significantly in the PD discrimination process (section 6.1.2). Perhaps this is because the subjects spend a long time holding the pen before or after the drawings task which increases the chance to obtain more useful information.

Time-series datasets include many useful features as this thesis described in section 3.1, and normally more good information leads to higher chance to classify patients and control individuals (section 6.1.4). drawing a 3D shape (cube) is harder than drawing a 2D shape (pentagon), especially for the PD patients which increases the chances of such an exam to distinguish patients from healthy individuals. Although we think that both RNN and CNN topologies were effective in this research, using RNN on subject's movement signals that have been taken during cube drawing exam has obtained better results than other approaches.

Our results concur with Pereira et al. (2017) results that the harder disease exam is more discriminative and effective to distinguish healthy subjects from individuals with PD. On the other hand, Although Pereira et al. (2016) found that 64x64 resolution may not represent the whole data well enough to be discriminated by CNNs, we found this resolution gives us the best results in terms of CNN-based experiments. Also, we agree that an imbalanced dataset affects negatively on the classification results as Pereira et al. (2016) noticed. And finally, our results using deep learning techniques exceed what Pereira et al. (2015) achieved using traditional machine learning algorithms.

## 6.2 Achievements

During this project, many predictive models were developed to differentiate between healthy people and people with PD. These models were developed within a variety of experiments using two deep learning algorithms (RNN and CNN) and two datasets (cube and pentagon). These models present different results and performance. For time series datasets, we investigated which dataset achieved better results and performance using two different experiments for each dataset. For imaging datasets, we carried out comparisons using 6 different experiments for each dataset to examine which one obtained the best results. Reaching this point enabled us to explore which dataset type is more effective as a basis for discrimination by statistically analysing the experiment's results distribution. At the same time, this analysis enabled us to investigate which deep learning technique produces the best results and performance for these datasets.

# Chapter 7. Conclusion & Future work

## 7.1 Project Problems

- **Imbalanced datasets:** cube and pentagon datasets are imbalanced as we described in chapter 3. This problem misled the classifier where our models classified all the test set as patients and this is bad behaviour. For CNNs, the solution was to balance our datasets using augmentation as we described in section 4.1.3. For RNNs, the solution was to increase the number of neurons in the LSTM layer and reduce the batch size.

- **Variation of lengths in time-series sequences with the dataset:** normally deep learning networks require a fixed size of input and samples with different lengths do not work in such networks. Our solution was zero-padding before each sample until reaching the longest sample in a dataset.

- **Computational resources:** using deep learning implies a high computational cost. Therefore, a machine with a strong specification is needed to train the models. We trained CNN models using free GPU google Collaboratory and this reduced the training time by 10 times. In addition, we used images normalization which improved the classifier performance and reduced the number of epochs required to learn and reach very good accuracy as we described in section 4.1.3. We trained RNN models using our machine CPU, which took a long time to finish.

## 7.2 Limitations

- As we described in chapter 3, the pentagon dataset includes 4 tries for each subject. In our work, we consider each try as a separate subject. However, PD affects the body asymmetrically. This means that the patient's group might include patients with PD symptoms in one hand and no symptoms in the other hand (i.e. this hand draws like a healthy subject hand). Therefore, these tries mislead the classifier as the model will learn these data to distinguish between PD and healthy subjects, but the data under patient group is the same as data from healthy hands.

- We used the validation set only to investigate the model performance during the training and this reduced the number of samples in the training set.

- RNN training is too slow and this not flexible in practice work.

- Disconnecting and resources exhaustion: working with cloud services like google Collaboratory causes many problems like disconnecting suddenly. And because it is shareable service by the world zones, this leads to resources exhaustion error many times.

### 7.3 Future work:

To cope with the limitations described above our future work would be the following:

- As we described in section 7.2 about the patients trying, the metadata can be used for each hand, since the UPDRS scores show how each hand is affected. In this way, we will make sure that we use the data from the afflicted hand, so our classifier will learn properly.

- The validation set can be used for early stopping. Therefore, we will avoid the overfitting and reduce the classification error remarkably (Prechelt, 1998).

- In order to speed up the RNN training, it can be used Truncated Backpropagation Through Time TBPTT. In TBPTT, each time-series sequence is divided into chunks, then we run forward and backward passes through these chunks instead of the whole sequence. This speeds up RNN performance and it does not suffer from exploding gradient problem (Brownlee, 2017). as well as, we can reduce the training time using Amazon web services AWS.

- Using meta-heuristic (optimization algorithms) such as the Evolutionary algorithm to choose deep learning hyper-parameters and configurations to provide the best results and performance.

- Investigate deeper networks or complex topologies like Long-term Recurrent Convolutional Network (LRCN) which uses the extracted features from CNN layers as LSTM input.

From the perspective of expanding our work, our future work would focus on:-

- Investigate whether the developed models can give information about disease staging. e.g. Can the models differentiate between patients have cognitive and patients without?.

- Examine if we can extract knowledge from the trained models and understand the basis of their discrimination, i.e. understanding the features that the models use it to classify PD patients.

## 7.4 Conclusion

In this thesis, we investigated Parkinson's Disease recognition by means of deep learning. We explored two deep learning algorithms, RNN and CNN, on two different PD datasets, Cube and Pentagon, including two different datasets types, time series and images.

The experimental workflows were designed to analyse the performance of the deep learning models on a variety of modified datasets in order to choose the best triple: deep learning technique, dataset and dataset type. These workflows start with some data processing passing through the classification task and reaching by collect the results using cross-validation. After that analyse the experiments performance statistically then evaluating the models for the experiment with the best results.

Our results demonstrate that deep learning algorithms are well behaved in the classification of both motor signals and images which are extracted during PD exams. In addition, the results show that the harder exam has the better chance to recognize the difference between the patients and the healthy individuals.

# Appendix

The appendix I and II are provided by a separate file